\documentclass[10pt,twocolumn,letterpaper]{article}

\usepackage[pagenumbers]{wacv} 

\usepackage{graphicx}
\usepackage{amsmath}
\usepackage{amssymb}
\usepackage{booktabs}
\usepackage{comment}
\usepackage{makecell}
\usepackage{multirow}
\usepackage{subcaption}
\usepackage{listings}
\usepackage{pifont}
\usepackage{microtype} 
\usepackage[normalem]{ulem}
\usepackage[accsupp]{axessibility}
\useunder{\uline}{\ul}{}

\usepackage{tikz}
\usetikzlibrary{arrows, backgrounds, calc, decorations, calligraphy, external, fit, positioning, shapes, spy}

\usepackage[pagebackref,breaklinks,colorlinks]{hyperref}

\usepackage[capitalize]{cleveref}
\crefname{section}{Sec.}{Secs.}
\Crefname{section}{Section}{Sections}
\Crefname{table}{Table}{Tables}
\crefname{table}{Tab.}{Tabs.}

\def\wacvPaperID{66} 
\def\confName{WACV}
\def\confYear{2024}

\begin{document}

\title{Attention Modules Improve Image-Level Anomaly Detection for Industrial Inspection: A DifferNet Case Study}

\author{André Luiz Vieira e Silva$^{1,*}$
\and
Francisco Simões$^{1,3}$
\and
Danny Kowerko$^{2}$
\and
Tobias Schlosser$^{2}$
\and
Felipe Battisti$^{1}$
\and
Veronica Teichrieb$^{1}$
\and
$^{1}$ Voxar Labs, Centro de Informática, Universidade Federal de Pernambuco, Brazil\\
$^{2}$ Junior Professorship of Media Computing, Chemnitz University of Technology, Germany\\
$^{3}$ Visual Computing Lab, DC, Universidade Federal Rural de Pernambuco, Brazil\\
$^{*}$ {\tt\small albvs@cin.ufpe.br}
}

\maketitle

\begin{abstract}
Within (semi-)automated visual industrial inspection, learning-based approaches for assessing visual defects, including deep neural networks, enable the processing of otherwise small defect patterns in pixel size on high-resolution imagery. The emergence of these often rarely occurring defect patterns explains the general need for labeled data corpora. To alleviate this issue and advance the current state of the art in unsupervised visual inspection, this work proposes a DifferNet-based solution enhanced with attention modules: AttentDifferNet. It improves image-level detection and classification capabilities on three visual anomaly detection datasets for industrial inspection: InsPLAD-fault, MVTec AD, and Semiconductor Wafer. In comparison to the state of the art, AttentDifferNet achieves improved results, which are, in turn, highlighted throughout our quali-quantitative study. Our quantitative evaluation shows an average improvement~-- compared to DifferNet~-- of $\mathit{1.77 \pm 0.25}$ percentage points in overall AUROC considering all three datasets, reaching SOTA results in InsPLAD-fault, an industrial inspection in-the-wild dataset. As our variants to AttentDifferNet show great prospects in the context of currently investigated approaches, a baseline is formulated, emphasizing the importance of attention for industrial anomaly detection both in the wild and in controlled environments.
\end{abstract}

\section{Introduction}

The automation of visual defect inspection can reduce inspection costs and security risks in multiple industries. However, industries such as manufacturing \cite{bergmann2019mvtec, rudolph2021differnet, yu2021fastflow}, healthcare \cite{schlegl2019fanogan,han2021madgan}, security \cite{akcay2019skipganomaly}, video surveillance \cite{roth2022patchcore, defard2021padim}, and power delivery \cite{ge2021anomaly, vieiraesilva2021stnplad} often suffer with the scarcity of defective samples to train deep learning methods due to their rare occurrence and their high financial and social impact. Those factors severely hamper the usage of fully supervised machine learning approaches while increasing the popularity of un-/semi-supervised anomaly detection methods \cite{liu2023deep}.

Anomaly detection methods often rely on normal/flawless samples during model training. They extract unique information from those samples, e.g., data distributions, whereby during test time, they can discriminate between flawless and anomalous samples. The recent MVTec AD \cite{bergmann2021mvtec} dataset for anomaly detection fostered new research on this topic, such as anomaly detection methods based on normalizing flows, which are a class of machine learning models that are used for density estimation. This approach has become popular since it can model complex probability distributions using simpler ones, e.g., normal distributions. 

Although most recent anomaly detection methods use MVTec AD as their primary dataset \cite{rudolph2022csflow, rudolph2021differnet, roth2022patchcore, yu2021fastflow, gudovskiy2022cflowad, defard2021padim, tien2023revisiting}, it only presents limited challenges focused on the manufacturing industry. The components are captured under a controlled environment with constant background, lighting, object scale, perspective, and image resolution. More recently, some datasets address this issue, such as the AeBAD dataset \cite{zhang2023industrial}, which provides the diversity of domains within the same data category, and the MVTec LOCO-AD \cite{bergmann2022Beyond}, which evaluates logical constraints in anomaly detection.

Anomaly detection for industrial inspection in the wild, e.g., for power line inspection, is an open problem due to the lack of public datasets and the associated computer vision challenges, which include the variation of perspective, scale, orientation, lighting, background, and resolution, as well as cluttering and projection deformations due to multiple camera angles.

Attention modules can boost the representational power of artificial neural networks by improving spatial and/or channel encodings. In other words, they highlight relevant information from foreground objects while concealing the background and other less relevant image regions and objects. Attention modules can be easily integrated into most CNN-based methods, improving feature extraction quality without compromising computational performance.

This work studies the usage of attention modules on DifferNet, a modern anomaly detection method based on normalizing flows. The main findings of this work are:

\begin{itemize}
    \item The new attention-based DifferNet, AttentDifferNet, is superior to the standard DifferNet on all objects from three anomaly detection datasets, each dataset from a distinct industrial inspection domain;
    \item AttentDifferNet achieves state-of-the-art performance on InsPLAD-fault, a dataset for image-level industrial anomaly detection in the wild;
    \item AttentDifferNet is qualitatively superior to DifferNet;
    \item A straightforward coupling of popular attention modules to modern feature-embedding-based unsupervised anomaly detection.
\end{itemize}

\section{Related Work}

Conventionally, classical (semi-)automated visual inspection approaches encompassed the following \cite{Huang2015, schlosser2022improving}: projection-based (principal component analysis (PCA), linear discriminant (LDA), or independent component analysis (ICA) based approaches), filter-based (spectral estimation and transformation based approaches such as discrete cosine (DCT), Fourier (FT), and wavelet transform), as well as hybrid approaches \cite{sreenivasan1993automated, zhang1999development, Chen2000, tobin2001integrated}. With the rise in computational power and the availability of large labeled data corpora, learning-based approaches, including support vector machines (SVM) and artificial neural networks (ANN) such as multilayer perceptrons (MLP), have long since taken over the competition over the last decades \cite{Chen2000, liu2002wafer, Huang2007, Xie2014}. Today, deep neural networks (DNN) such as convolutional neural networks (CNN) \cite{Nakazawa2018, Cheon2019, OLeary2020, schlosser2022improving} and vision transformers (ViT) are considered the vanguards of human-like performance in terms of detection and classification capabilities.

Current works apply attention mechanisms to classic image-level anomaly detection methods, encompassing approaches based on CNNs \cite{takimoto2022anomaly, song2023research, sun2022scale} and generative adversarial networks (GAN) \cite{ge2021anomaly}. However, due to the proposition of public datasets for anomaly detection such as MVTec AD and Magnetic Tiles Defects (MTD) \cite{huang2020surface}, new classes of anomaly detection methods have been proposed. Those modern methods are the current state of the art considering the benchmarks built from MVTec AD and MTD. Currently, a popular approach is to propose methods that benefit from extracted feature embeddings at image and pixel levels. Two recent techniques are distribution mapping through normalizing flows \cite{rudolph2022csflow, gudovskiy2022cflowad, rudolph2021differnet, yu2021fastflow} and feature memory banks \cite{defard2021padim, roth2022patchcore, cohen2020spade}.

Normalizing flows are commonly used for density estimation. It uses a series of invertible mathematical transformations to turn samples from the base distribution into samples from the target distribution. Thus, they are used to learn the underlying probability density function of the normal data. Finally, any data point with a low likelihood is considered an anomaly under the learned model.
A feature memory bank is another recent approach in which the extracted feature embeddings are stored in a memory bank. Each method uses a different approach to how the features are grouped and how they relate to each other. In the testing phase, when an image is presented, its features are extracted and compared to the ones in the memory bank. Based on the similarity of the features, the tested image is classified as normal or anomalous.
Both approaches similarly use a backbone CNN to extract features. In that sense, using attention mechanisms in that phase may assist the network.

Recently, multiple attention modules have been proposed to improve the expression ability of CNNs in visual inspection \cite{ristea2022selfsupervised, bhattacharya2021interleaved}. To improve the spatial and/or channel encoding, multiple image-level anomaly detection methods apply attention mechanisms \cite{takimoto2022anomaly, song2023research, sun2022scale, ge2021anomaly}. On a similar path, in \cite{yan2022cainnflow}, the authors proposed to apply attention blocks during the normalizing flow step, which can lead to complex modifications due to their mathematically invertible nature.

Two of the most popular are the Squeeze-and-Excitation Networks (SENet) \cite{hu2018senet} and the Convolutional Block Attention Module (CBAM) \cite{woo2018cbam}. SENet can adaptively recalibrate channel-wise response with global contextual information by signals aggregated from feature maps. CBAM introduces channel and spatial attention to generate weights of different channels and locations, highlighting the location and class information. Other popular modular attention mechanisms that further develop the idea of channel and spatial attention are ECANet \cite{wang2020ecanet} and SA-Net \cite{zhang2021sanet}. They may also be integrated into many architectures. 

We propose a new method using modular attention mechanisms on an anomaly detection method based on normalizing flows through simple modifications. The hypothesis in this work is that the usage of attention modules helps a state-of-the-art anomaly detection method to focus on the analyzed foreground object while overcoming background interference and other distractions for industrial anomaly detection in the wild. For this purpose, it is also crucial that the proposed method's performance does not deteriorate in controlled scenarios.

\section{Methods}

In this section, we explain how DifferNet and the applied attention modules work and how we combine them to develop our proposed method, AttentDifferNet. 

\subsection{DifferNet}

DifferNet \cite{rudolph2021differnet} is a recent method for unsupervised image-based anomaly detection. Preliminary experiments indicated that DifferNet's performance for anomaly detection in the wild was superior to other similar methods, including even to more recent ones. These results can be seen in Table~\ref{tab:insplad} in Section~\ref{sec:exp}. Therefore, DifferNet was our first choice to be adapted with attention modules to enhance the focus on objects of interest in the wild.

Regarding the method itself, DifferNet combines convolutional neural networks with normalizing flows. The CNN in DifferNet is an AlexNet \cite{krizhevsky2017alexnet}, which works as a backbone for feature embedding extraction. It takes the training images to generate descriptive features of flawless images. The features are then mapped to a latent space using a normalizing flow model. It is possible to calculate the likelihood of image samples from this latent space, whereas anomalous images should present a lower likelihood than the flawless samples present in the training process. Because of this, the training goal is to find parameters that maximize the likelihood of extracted features in the latent space.

\subsection{Attention Modules}

Squeeze-and-Excitation Networks \cite{hu2018senet} and Convolutional Block Attention Modules \cite{woo2018cbam} are two well-known architectural unit attention modules with similar goals: increase the representational power of CNNs by selectively emphasizing important features while suppressing irrelevant ones. 

\subsubsection{Squeeze-and-Excitation Networks (SENet)}

The SENet \cite{hu2018senet} method consists of three submodules: the Squeeze Module, the Excitation Module, and the Scale Module, also shown in Figure~\ref{fig:senet}. The Squeeze Module focuses on adapting feature maps for optimal channel attention. It utilizes a feature descriptor such as pooling to reduce the spatial dimensions of feature maps to a single value, resulting in the attention being adaptive to each channel. By decomposing each feature map, the computational complexity is reduced. The global average pool descriptor (GAP) is chosen as the feature descriptor, which calculates the average of all pixels within the feature map.

The Excitation Module employs an MLP bottleneck structure. It reduces the input space to a smaller dimension using a reduction factor and then expands it back to the original dimensionality. The MLP operates on the compressed space and maintains the $(C\times1\times1)$ shape throughout the module. Finally, in the Scale Module, the ``excited'' tensor undergoes a sigmoid activation to scale the values to a range of $[0, 1]$. Subsequently, the output is multiplied element-wise with the input tensor using broadcasted multiplication. Each channel~/ feature map in the input tensor is scaled by its corresponding learned weight from the MLP in the Excitation Module.

\subsubsection{Convolutional Block Attention Modules (CBAM)}

\begin{figure}[t]
    \centering
    \begin{subfigure}{\linewidth}
        \centering
        \includegraphics[width=.75\linewidth]{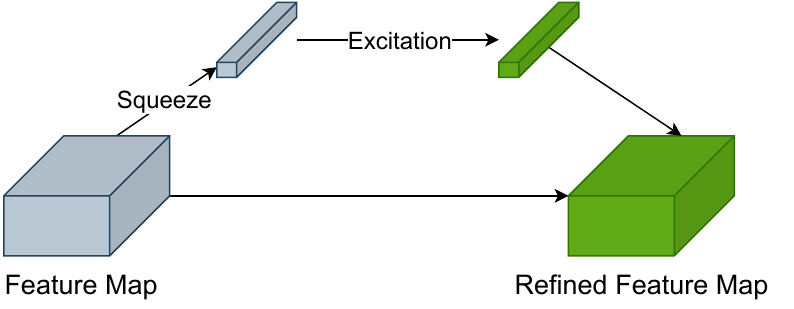}
        \caption{SENet module architecture representation.}
        \label{fig:senet}
    \end{subfigure}
    
    \bigskip
    
    \begin{subfigure}{\linewidth}
    \centering
        \includegraphics[width=.75\linewidth]{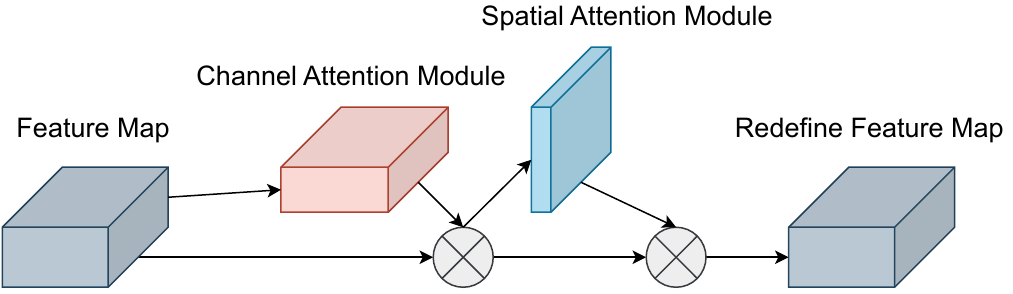}
        \caption{CBAM module architecture representation.}
        \label{fig:cbam}
    \end{subfigure}
    \caption{Architecture of applied attention modules.}
    \label{fig:att_mods}
\end{figure}

CBAM \cite{woo2018cbam} utilizes a combination of two submodules, the channel attention module and the spatial attention module displayed in Figure~\ref{fig:cbam}. The first one creates an attention map using the inter-channel connections of the features. The method extracts spatial context descriptors by combining average pooling and max-pooling operations on a given feature map. Then, a shared network processes these descriptors, resulting in a channel attention map.

To generate a spatial attention map highlighting informative regions, CBAM employs a spatial attention module that captures inter-spatial relationships within the features. Unlike channel attention, which focuses on the importance of different feature channels, spatial attention determines the location of informative parts. First, to compute the spatial attention, average pooling and max-pooling operations are applied along the channel axis. The resulting feature maps are concatenated. This concatenated feature descriptor is fed through a convolutional layer, resulting in a spatial attention map indicating regions to emphasize or suppress.

\subsection{AttentDifferNet}

DifferNet was conceived to detect defects in objects from images captured in a controlled context such as objects from an industrial production line. To adapt it to overcome the challenges of object inspection in the wild, modular attention-based mechanisms were added to its backbone architecture. This allows the backbone network to focus on foreground elements to generate more relevant feature embeddings of the image with the inspected object. In this work, two architectures are experimented with, one using SENet and one using CBAM, as multiple works report significant performance increases by adding them to the pipeline while they can be easily coupled into CNNs. Furthermore, they are arguably two of the most popular attention modules.

\begin{figure}[t]
    \centering
    \includegraphics[width=\linewidth]{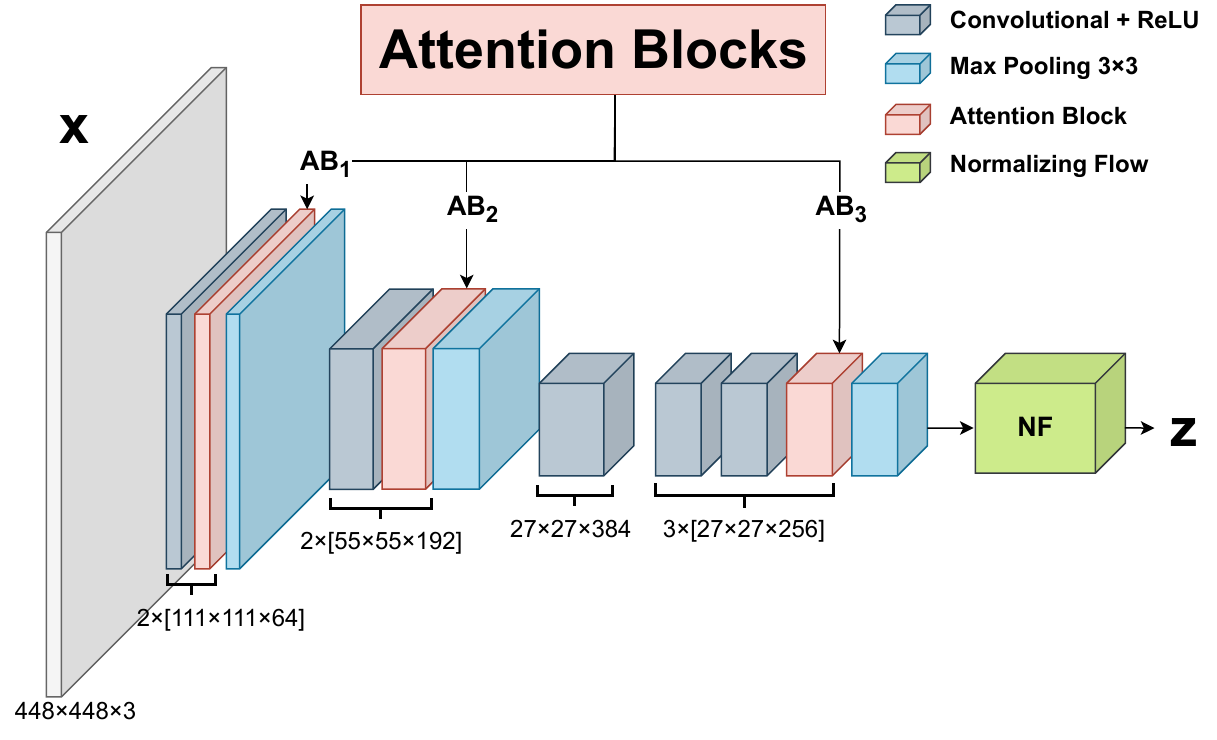}
    \caption{Proposed AttentDifferNet architecture.}
    \label{fig:attentdiffernet}
\end{figure}

Figure~\ref{fig:attentdiffernet} shows our proposed architecture. The attention block's role changes according to the depth in which it is placed within the neural network. In the first few layers, it learns to highlight lower-level, class-agnostic features. In the deeper layers, it becomes more specialized, responding to different inputs in a class-specific manner. Therefore, our proposed architecture leverages the advantages of attention blocks throughout the entire network.

Similarly to DifferNet, AttentDifferNet works by extracting image feature embeddings $y \in Y$ from an anomaly-free input image $x \in X$ and estimating these feature embeddings' density. The $f_{\textit{FE}}: X \rightarrow Y$ map is done by a pretrained feature extractor, which is a modified version of AlexNet with three attention blocks: AB\textsubscript{1}, AB\textsubscript{2}, and AB\textsubscript{3}, positioned as shown in Figure~\ref{fig:attentdiffernet}. To estimate $p_Y(y)$, the probability density function, it is first necessary to map from $Y$ into a latent space $Z$, the latter having a known and well-defined $p_Z(z)$. This mapping is achieved by applying a normalizing flow $f_{\textit{NF}}: Y \rightarrow Z$. Finally, the likelihood of a given input image is calculated directly from $p_Z(z)$. A high likelihood indicates that a set of image features is from the distribution, while a low one indicates otherwise. Thus, the latter should represent an anomalous sample.

\subsection{Attention Modules in Other Methods}

To show how attention modules impact other AD methods, we added them to SOTA AD methods according to MVTec AD: ReverseDistillation++ \cite{tien2023revisiting} and FastFlow \cite{yu2021fastflow}. To obtain a fair comparison, we select ResNet-18 as their backbone network, which is somewhat deeper than AlexNet. To combine ResNet-18 with attention modules, we follow the SENet authors' orientation by placing the SENet block in the non-identity branch of a residual module, resulting in the Squeeze and Excitation operations both to act before the summation with the identity branch, producing SE-RN18 and CBAM-RN18.

\section{Experiments}
\label{sec:exp}

\subsection{Datasets}

Here, we present the three industrial inspection datasets used in our experiments: one in the wild, which is remotely sensed using a drone, and two in controlled scenarios.

\begin{figure*}
\scriptsize
    \centering
    \setlength{\tabcolsep}{1pt}
    \begin{tabular}{cc@{\hskip 8pt}cc@{\hskip 8pt}cc@{\hskip 8pt}cc@{\hskip 8pt}cc}
        \multicolumn{2}{c}{Glass Insulator} & \multicolumn{2}{c}{Lightning Rod Suspension} & \multicolumn{2}{c}{Polymer Insulator Upper Shackle} & \multicolumn{2}{c}{Vari-grip} & \multicolumn{2}{c}{Yoke Suspension}\\
        \noalign{\vskip 2pt}     
        \textcolor{green}{\frame{\includegraphics[width=0.19\columnwidth]{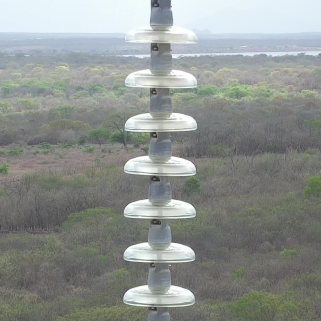}}}  &
        \textcolor{green}{\frame{\includegraphics[width=0.19\columnwidth]{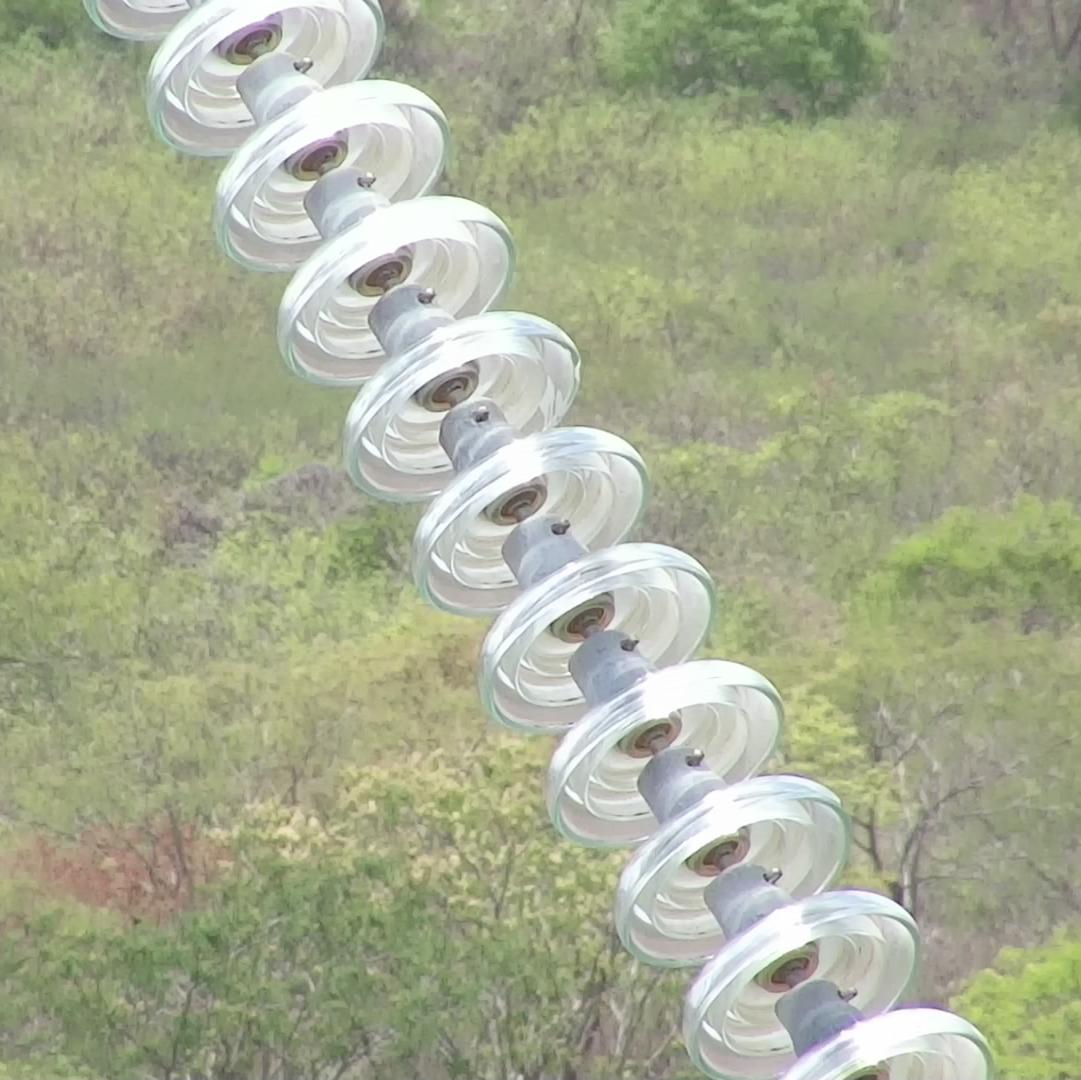}}} &
        \textcolor{green}{\frame{\includegraphics[width=0.19\columnwidth]{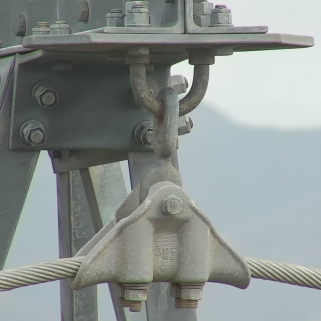}}}  &
        \textcolor{green}{\frame{\includegraphics[width=0.19\columnwidth]{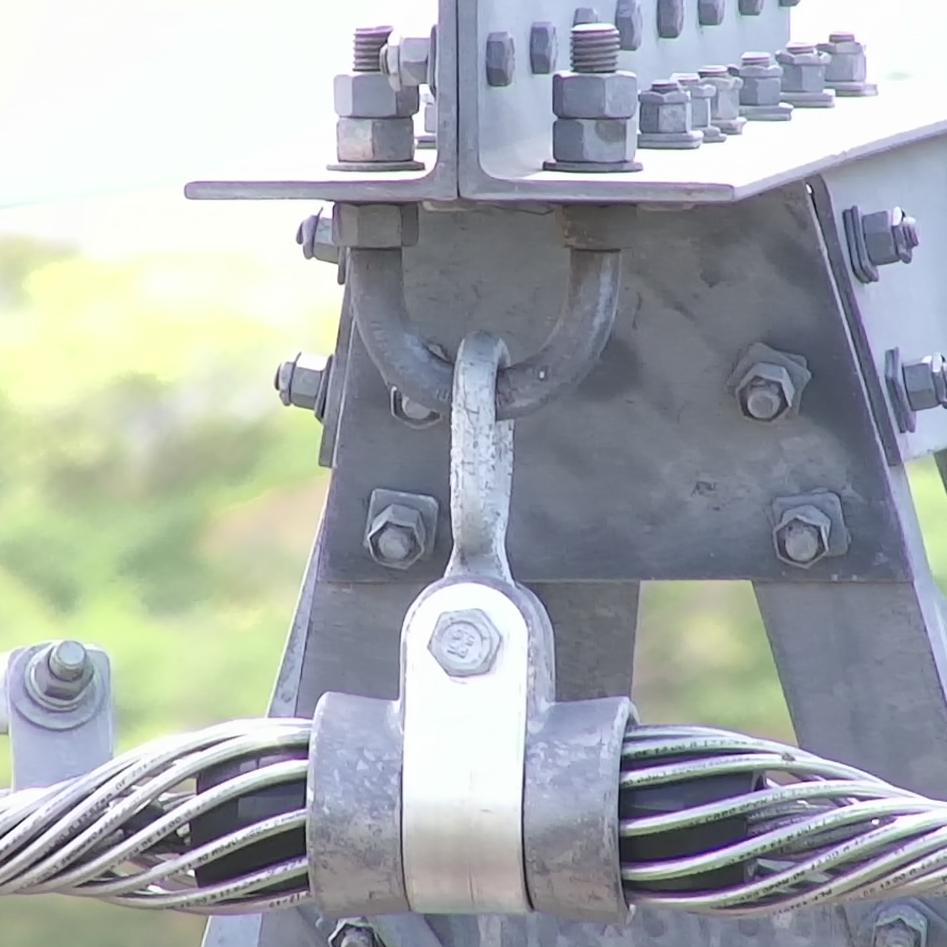}}} &
        \textcolor{green}{\frame{\includegraphics[width=0.19\columnwidth]{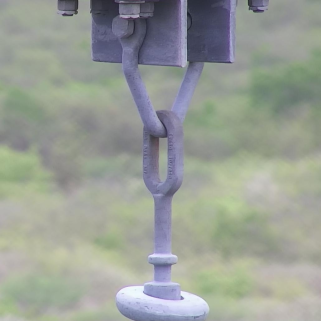}}} &
        \textcolor{green}{\frame{\includegraphics[width=0.19\columnwidth]{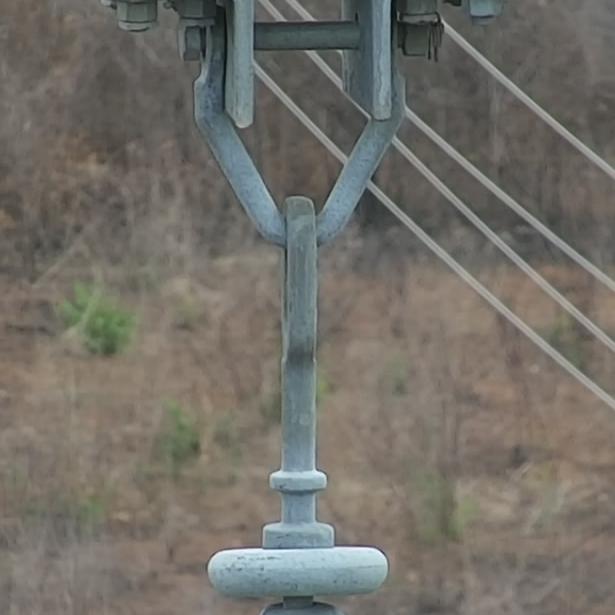}}} &
        \textcolor{green}{\frame{\includegraphics[width=0.19\columnwidth]{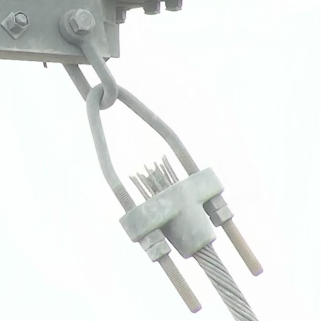}}} &
        \textcolor{green}{\frame{\includegraphics[width=0.19\columnwidth]{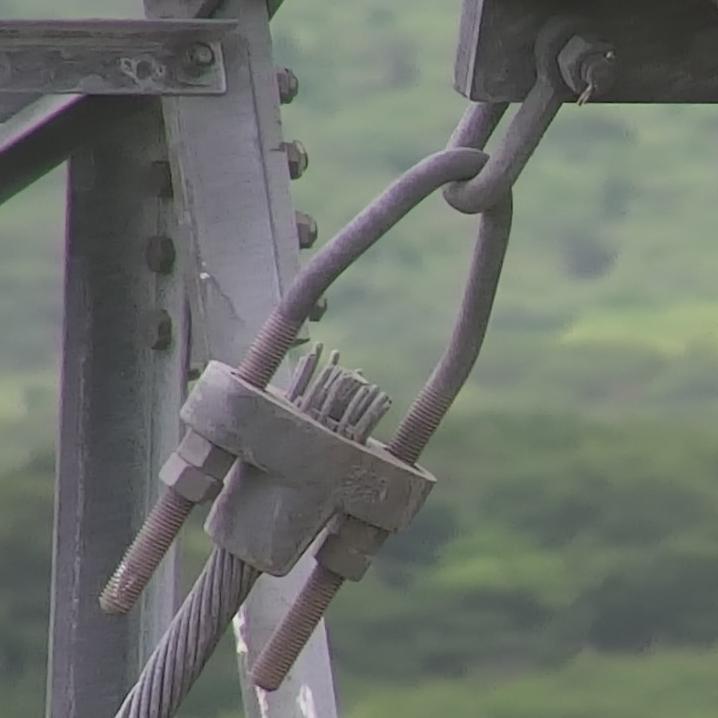}}} &
        \textcolor{green}{\frame{\includegraphics[width=0.19\columnwidth]{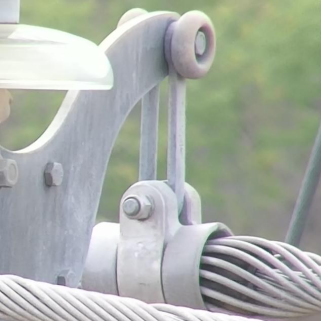}}} &
        \textcolor{green}{\frame{\includegraphics[width=0.19\columnwidth]{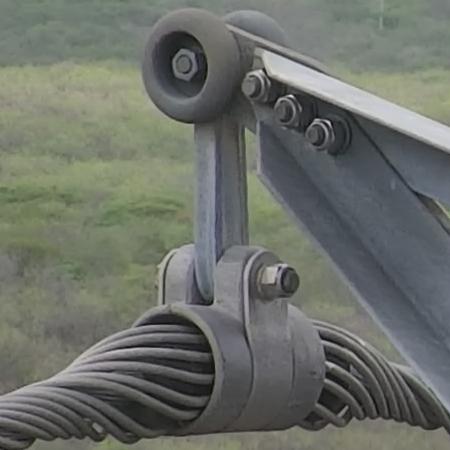}}} \\
        \textcolor{red}{\frame{\includegraphics[width=0.19\columnwidth]{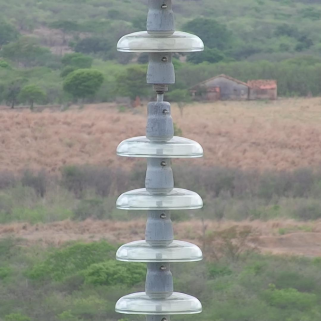}}}  &
        \textcolor{red}{\frame{\includegraphics[width=0.19\columnwidth]{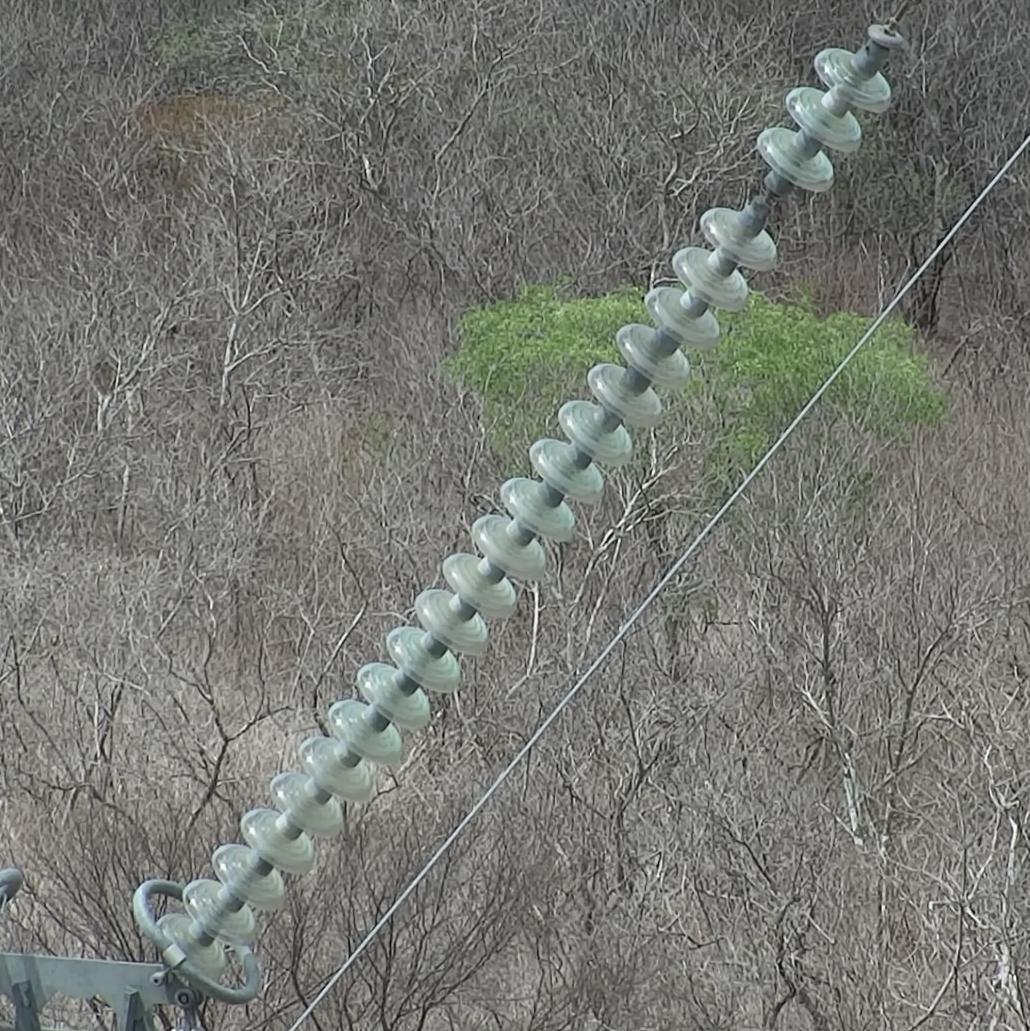}}} &
        \textcolor{red}{\frame{\includegraphics[width=0.19\columnwidth]{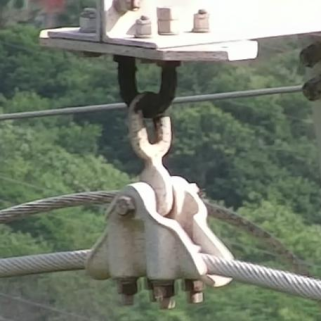}}}  &
        \textcolor{red}{\frame{\includegraphics[width=0.19\columnwidth]{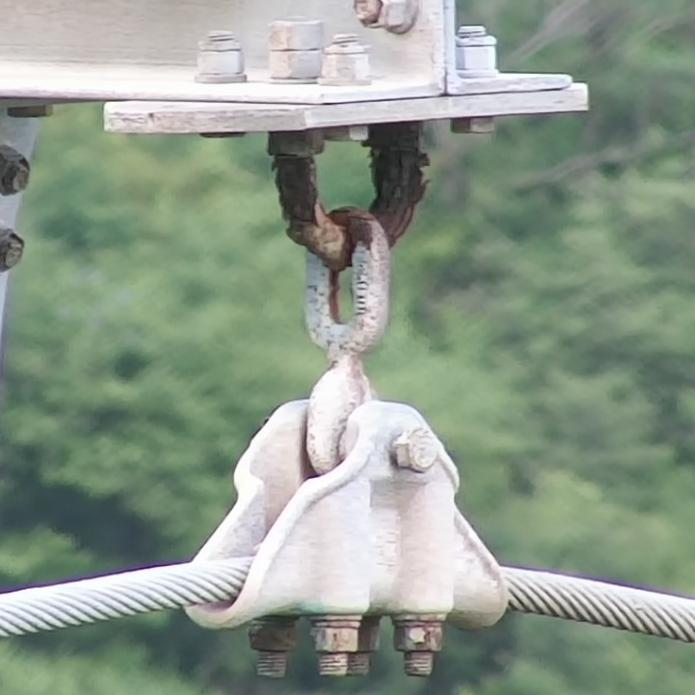}}} &
        \textcolor{red}{\frame{\includegraphics[width=0.19\columnwidth]{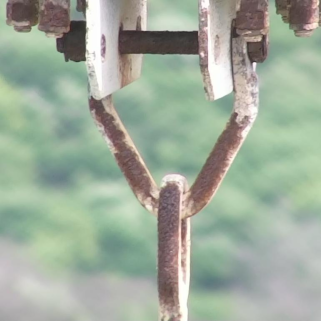}}} &
        \textcolor{red}{\frame{\includegraphics[width=0.19\columnwidth]{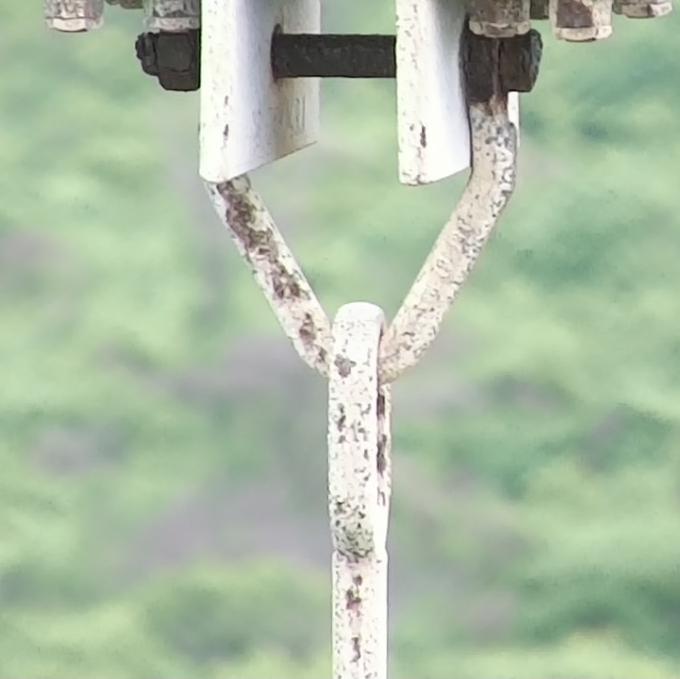}}} &
        \textcolor{red}{\frame{\includegraphics[width=0.19\columnwidth]{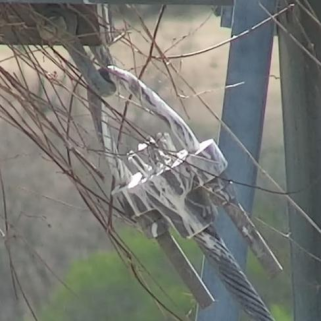}}} &
        \textcolor{red}{\frame{\includegraphics[width=0.19\columnwidth]{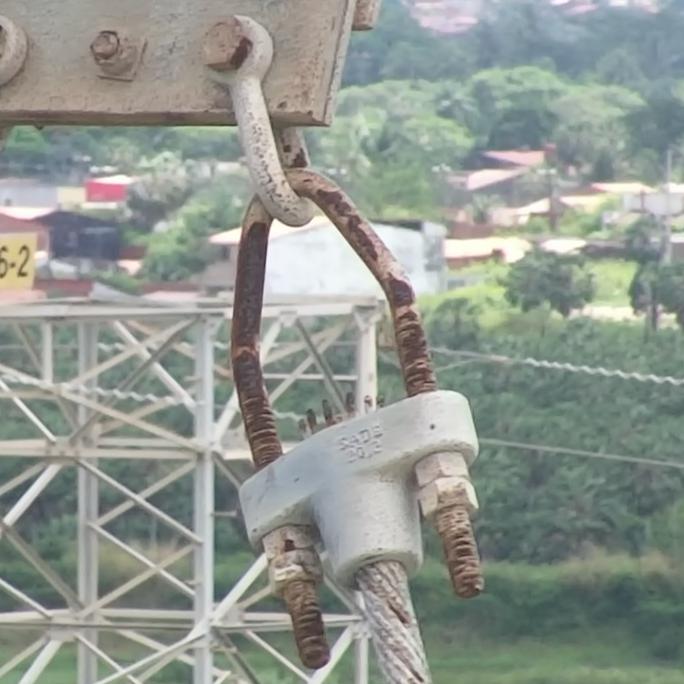}}} &
        \textcolor{red}{\frame{\includegraphics[width=0.19\columnwidth]{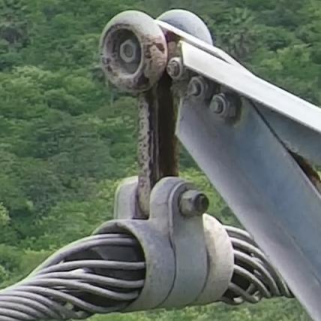}}} &
        \textcolor{red}{\frame{\includegraphics[width=0.19\columnwidth]{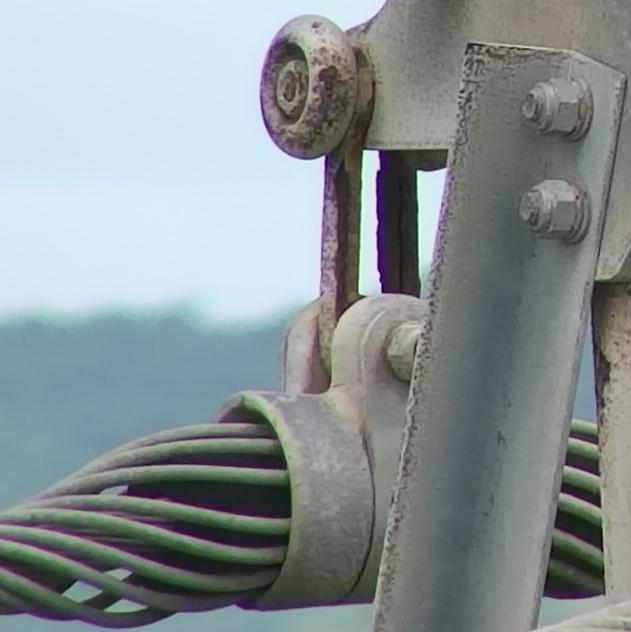}}} \\

    \end{tabular}

    \caption{InsPLAD-fault dataset overview. The first row shows flawless samples (green frames), while the second row shows defective ones (red frames).
    }
    \label{fig:insplad}
\end{figure*}

\textbf{InsPLAD}\footnote{\url{https://github.com/andreluizbvs/InsPLAD} (see \texttt{unsupervised\_anomaly\_detection.zip})} \cite{vieiraesilva2023insplad} is a power line asset inspection in-the-wild dataset that offers multiple computer vision challenges, one being anomaly detection in power line components called InsPLAD-fault. Its data are real-world unmanned aerial vehicle (UAV) images of operating power line transmission towers. It contains five power line object categories with one or two types of anomalies for each class, resulting in $11\,662$ images, of which $402$ are samples of defective objects annotated on image level. Since they are real-world defects, none of the faults have been fabricated or generated manually. Table~\ref{tab:inspladdes} shows the InsPLAD-fault properties for the anomaly detection task, whereas Figure~\ref{fig:insplad} depicts a flawless and a defective sample for each of the five power line object classes.

\begin{table}
  \centering
  \resizebox{\linewidth}{!}{%
\begin{tabular}{@{\extracolsep{\fill}}lccc@{\extracolsep{\fill}}}
\toprule
\multirow{4}{*}{Asset category} & \multicolumn{3}{c}{Anomaly detection}     \\ \cmidrule(lr){2-4}  
                                & \multicolumn{1}{c}{Train} & \multicolumn{2}{c}{Test}      \\ \cmidrule(lr){2-2} \cmidrule(lr){3-4}
                                & Flawless        & Flawless        & Anomalous         \\ 
\midrule
Glass Insulator             & 2298         & 581         & 90           \\
Lightning Rod Suspension    & 462          & 117         & 50           \\
\makecell[l]{Polymer Insulator Upper Shackle}    & 935          & 235         & 102          \\
Vari-grip                   & 477          & 114         & 63/48        \\ 
Yoke Suspension             & 4834         & 1207        & 49           \\
    \bottomrule
  \end{tabular}
  }
  \caption{InsPLAD-fault anomaly detection dataset description. Glass Insulator anomalies are missing caps, while the remaining are corrosion-related. Vari-grip has two types: bird nest~/ corrosion.}
  \label{tab:inspladdes}
\end{table}

\begin{figure}[tbp]
	\newcommand{\xscale}{2}
	\newcommand{\yscale}{1.75}
	\newcommand{\width}{0.2\columnwidth}
	\newcommand{\tcgf}[1]{\textcolor{green}{\frame{#1}}}
	\newcommand{\tcrf}[1]{\textcolor{red}{\frame{#1}}}

	\centering

	\resizebox{\columnwidth}{!}{\begin{tikzpicture}[xscale=\xscale, yscale=\yscale]
		\begin{scope}
			\node at (0, 0) {Type 1};
			\node at (1, 0) {Type 2};
			\node at (2, 0) {Type 3};
			\node at (3, 0) {Type 4};
			\node at (4, 0) {Type 5};
		\end{scope}

		\begin{scope}[yshift=-0.75cm]
			\node at (0, 0) {\tcgf{\includegraphics[width=\width]{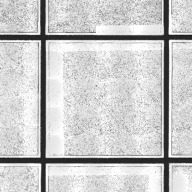}}};
			\node at (1, 0) {\tcgf{\includegraphics[width=\width]{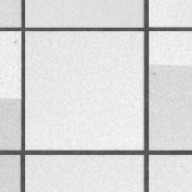}}};
			\node at (2, 0) {\tcgf{\includegraphics[width=\width]{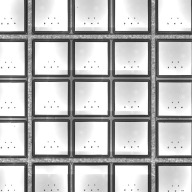}}};
			\node at (3, 0) {\tcgf{\includegraphics[width=\width]{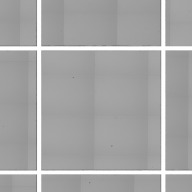}}};
			\node at (4, 0) {\tcgf{\includegraphics[width=\width]{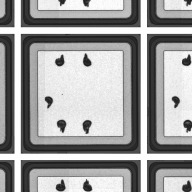}}};
		\end{scope}

		\begin{scope}[yshift=-1.8cm]
			\node at (0, 0) {\tcrf{\includegraphics[width=\width]{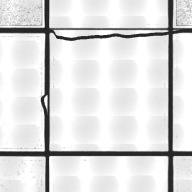}}};
			\node at (1, 0) {\tcrf{\includegraphics[width=\width]{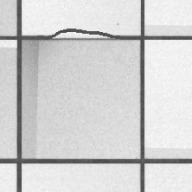}}};
			\node at (2, 0) {\tcrf{\includegraphics[width=\width]{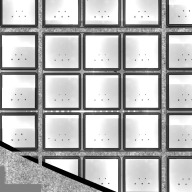}}};
			\node at (3, 0) {\tcrf{\includegraphics[width=\width]{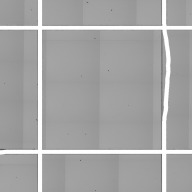}}};
			\node at (4, 0) {\tcrf{\includegraphics[width=\width]{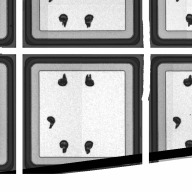}}};
		\end{scope}

		\begin{scope}[yshift=-2.95cm]
			\node at (0, 0) {\tcgf{\includegraphics[width=\width, trim={2cm 4cm 2cm 0}, clip]{{images/Wafer_synthesized/type_1/dataset_21_wafer_01_x_-000.5_y_+002.5}.jpg}}};
			\node at (1, 0) {\tcgf{\includegraphics[width=\width, trim={2cm 4cm 2cm 0}, clip]{{images/Wafer_synthesized/type_2/dataset_25_wafer_01_x_-003.5_y_-028.5}.jpg}}};
			\node at (2, 0) {\tcgf{\includegraphics[width=\width, trim={2cm 4cm 2cm 0}, clip]{{images/Wafer_synthesized/type_3/dataset_26_wafer_01_x_+001.5_y_+006.5}.jpg}}};
			\node at (3, 0) {\tcgf{\includegraphics[width=\width, trim={2cm 4cm 2cm 0}, clip]{{images/Wafer_synthesized/type_4/dataset_27_wafer_01_x_-006.5_y_+003.5}.jpg}}};
			\node at (4, 0) {\tcgf{\includegraphics[width=\width, trim={2cm 4cm 2cm 0}, clip]{{images/Wafer_synthesized/type_5/dataset_28_wafer_01_x_+002.5_y_-008.5}.jpg}}};
		\end{scope}

		\begin{scope}[yshift=-4cm]
			\node at (0, 0) {\tcrf{\includegraphics[width=\width, trim={2cm 4cm 2cm 0}, clip]{{images/Wafer_synthesized/type_1/dataset_21_wafer_01_x_-011.5_y_-004.5}.jpg}}};
			\node at (1, 0) {\tcrf{\includegraphics[width=\width, trim={2cm 4cm 2cm 0}, clip]{{images/Wafer_synthesized/type_2/dataset_25_wafer_01_x_-000.5_y_+020.5}.jpg}}};
			\node at (2, 0) {\tcrf{\includegraphics[width=\width, trim={2cm 0 2cm 4cm}, clip]{{images/Wafer_synthesized/type_3/dataset_26_wafer_01_x_-004.5_y_-010.5}.jpg}}};
			\node at (3, 0) {\tcrf{\includegraphics[width=\width, trim={4cm 2cm 0 2cm}, clip]{{images/Wafer_synthesized/type_4/dataset_27_wafer_01_x_-001.5_y_+002.5}.jpg}}};
			\node at (4, 0) {\tcrf{\includegraphics[width=\width, trim={2cm 0 2cm 4cm}, clip]{{images/Wafer_synthesized/type_5/dataset_28_wafer_01_x_+003.5_y_-029.5}.jpg}}};
		\end{scope}
	\end{tikzpicture}}

	\caption{Semiconductor Wafer dataset overview including flawless (green frames) and faulty (red frames) chips and streets per wafer type \cite{schlosser2022improving}. In order to protect the intellectual property of the wafer imagery, the shown examples are synthesized given the original imagery while retaining a close resemblance \cite{Friedrich2023}.}
	\label{figure:wafer_dataset}
\end{figure}

\textbf{MVTec AD} \cite{bergmann2019mvtec} is the most popular public dataset for unsupervised anomaly detection. It contains annotated data of objects and textures in controlled industrial scenarios at both image and pixel levels with and without anomalies. The anomalies are manually generated in an attempt to mimic real-world defects. It has ten objects and five textures categories, also shown in Table~\ref{tab:mvtecad}. There are two types of annotations: image-level, i.e., a normal or anomalous object in the image, and pixel-level, i.e., a normal or anomalous pixel in the image in the form of an image mask. In this work, we only use image-level annotations.

The \textbf{Semiconductor Wafer Dataset} \cite{schlosser2022improving} is a visual inspection wafer dataset for image classification (annotated in image-level), encompassing various wafers, chips, streets, and street segments. Wafer images were obtained from different real-world dicing manufacturers by scanning the wafers' chips after their cutting process. Figure~\ref{figure:wafer_dataset} shows a dataset overview in which the first two rows correspond to flawless and faulty samples of the chip category, while the last two to the street category. Also, each column shows a different type of wafer to ensure variability. During the creation of this dataset, it was also found that images containing faulty chips or streets occurred at a much lower frequency than those without defects. Unlike MVTec AD, it contains real faulty data, including defects such as ``spur'', ``break out'', ``overetch'', and ``scratch'' defect patterns \cite{Chen2000, liu2002wafer}.

\subsection{Implementation Details}

We used an NVIDIA RTX 2080 Ti GPU in a Linux environment to conduct our experiments. The convolutional layer weights are initialized with their respective AlexNet layers pretrained with ImageNet, while the attention block layers are trained from scratch. All models were trained for 100 epochs twice, for which the best AUROC results from those runs are reported. Other parameters and hyperparameters, such as input image size, batch size, and learning rate, are the same as reported in the original DifferNet work \cite{rudolph2021differnet}. Models were only trained with image-level annotations in all three datasets, i.e., no pixel-level masks were used.

\subsection{Quantitative Results}

The following sections present the quantitative analysis results. The analysis utilizes the area under the receiver operating characteristic curve (AUROC) as the performance metric. Notably, in every table, the underlined values represent the maximum AUROC achieved by any DifferNet implementations within each category, while bold values indicate the highest AUROC across all methods.

\subsubsection{InsPLAD}
\label{sec:inspladcomp}

Table~\ref{tab:insplad} details the results in the InsPLAD dataset. AttentDifferNet (SENet) consistently achieves the highest AUROC scores in every category, outperforming the baseline method, DifferNet. The most significant improvement is in the Glass Insulator category. It performs with an AUROC of 86.57\%, surpassing DifferNet's score of 82.81\%. Relevant enhancements can also be verified across every other category.

\begin{table*}[]
\centering
\resizebox{\textwidth}{!}{%
\begin{tabular}{cccccccccccccc}
\hline
\textbf{Category}               & DifferNet     & \makecell{AttentDifferNet \\(SENet)} & \makecell{AttentDifferNet \\(CBAM)}   & \makecell{FastFlow \\(RN18)}        & \makecell{FastFlow \\(SE-RN18)} & \makecell{FastFlow \\(CBAM-RN18)}       & \makecell{RD++ \\(RN18)} & \makecell{RD++ \\(SE-RN18)}   & \makecell{RD++ \\(CBAM-RN18)} & CS-Flow  & CFLOW-AD & PatchCore \\ \hline
\makecell[l]{Glass \\Insulator}                 & 82.81    & \underline{\textbf{86.57}}   & 81.03       & 70.16        & 73.21     & 73.01     & 86.03          & 86.54     & 86.21 & 85.73  & 82.22  & 78.44   \\ \\ 
\makecell[l]{Light. \\Rod Susp.}                & 99.08    & \underline{\textbf{99.62}}   & 99.33       & 82.02        & 77.60     & 80.03     & 97.06          & 97.69     & 97.54 & 96.60  & 95.52  & 85.11   \\ \\
\makecell[l]{Pol. Ins. \\Upper Shackle}         & 92.42    & \underline{\textbf{94.62}}   & 92.10       & 77.43        & 61.49     & 76.28     & 81.96          & 82.04     & 83.67 & 88.40  & 86.60  & 81.02   \\ \\
\makecell[l]{Vari-\\Grip}                       & 91.20    & \underline{93.52}            & 88.99       & 65.54        & 65.74     & 74.64     & \textbf{93.88} & 93.52     & 93.85 & 91.53  & 90.37  & 91.92   \\ \\
\makecell[l]{Yoke \\Suspension}                 & 96.77    & \underline{\textbf{97.38}}   & 96.86       & 71.48        & 73.64     & 75.68     & 91.42          & 91.80     & 92.46 & 90.70  & 83.87  & 58.06   \\ \hline
Avg AUROC                                       & 92.46    & \underline{\textbf{94.34}}   & 91.66       & 73.33        & 73.21     & 75.93     & 90.07          & 90.32     & 90.75 & 90.59  & 87.72  & 78.91   \\ \hline
\end{tabular}%
}
\caption{Comparison of area under ROC results in \% on the InsPLAD-fault dataset. Bold font indicates the best category result, while underlined values are the best between DifferNet variations. RN18 refers to the network backbone, ResNet-18.}
\label{tab:insplad}
\end{table*}

AttentDifferNet (CBAM) exhibits competitive performance in most categories, although slightly lower than AttentDifferNet (SENet). This indicates that incorporating attention blocks in the backbone network enhances or at least matches the DifferNet performance in this case. Compared to methods such as CS-Flow, PatchCore, FastFlow, CFLOW-AD, RD++, and their variations, both iterations of AttentDifferNet consistently demonstrate superior performance. AttentDifferNet (SENet) achieves the highest overall AUROC of 94.34\%, followed by DifferNet at 92.46\%.

Notably, the attention-enhanced FastFlow (CBAM-RN18) surpasses the standard FastFlow. Similarly, improvements are observed for the attention-based RD++ methods, with minor performance increases compared to the base RD++. These results highlight the role of attention mechanisms in AUROC scores considering this uncontrolled scenario.

\subsubsection{MVTec AD}

Table~\ref{tab:mvtecad} presents the experiments' results on the MVTec AD dataset. Among the DifferNet variations, AttentDifferNet (CBAM) consistently achieves the highest AUROC scores in most categories, outperforming the standard DifferNet in all of them. Furthermore, AttentDifferNet (SENet) also performs best among DifferNet variations in several categories, including Bottle, Cable, Capsule, Transistor, and Zipper. Both variations are overall superior to the basic DifferNet. 

\begin{table*}[]
\centering
\resizebox{\textwidth}{!}{%
\begin{tabular}{cccccccccccccccc}
\hline
\textbf{Category} & DifferNet  & \makecell{AttentDifferNet \\(SENet)} & \makecell{AttentDifferNet \\(CBAM)}  & \makecell{FastFlow \\(RN18)} & \makecell{FastFlow \\(SE-RN18)} & \makecell{FastFlow \\(CBAM-RN18)} & \makecell{RD++ \\(RN18)} & \makecell{RD++ \\(SE-RN18)} & \makecell{RD++ \\(CBAM-RN18)} & CS-Flow & CFLOW-AD & PatchCore \\ \hline
Bottle            & 99.00      & {\ul 99.84}           & 99.68                   & 97.74          & 98.43     & 98.26           & \textbf{100.00}  & \textbf{100.00}       & \textbf{100.00}   &  99.80         & \textbf{100.00}    & \textbf{100.00}    \\
Cable             & 95.90      & {\ul 98.43}           & 96.65                   & 96.94          & 91.00     & 96.06           & 99.06         & 99.34              & \textbf{99.57} & 99.10          & 97.59           & 99.50           \\
Capsule           & 86.90      & {\ul 93.86}           & 92.58                   & 98.28          & 98.17     & \textbf{98.50}  & 96.86         & 96.21              & 95.57          & 97.10          & 97.68           & 98.10           \\
Carpet            & 92.90      & 93.74                 & {\ul 95.18}             & 98.78          & 95.51     & 98.69           & 99.92         & 99.80              & 99.84          & \textbf{100.00}   & 98.73           & 98.70           \\
Grid              & 84.00      & \textbf{90.89}        & {\ul 91.23}             & 98.73          & 89.81     & 98.33           & 99.83         & \textbf{100.00}       & \textbf{100.00}   & 99.00          & 99.60           & 98.20           \\
Hazelnut          & 99.30      & \textbf{99.89}        & {\ul \textbf{100.00}}   & 96.17          & 98.42     & 96.02           & \textbf{100.00}  & \textbf{100.00}       & \textbf{100.00}   & 99.60          & 99.98           & \textbf{100.00}    \\
Leather           & 97.10      & 98.61                 & {\ul 99.32}             & 99.62          & 99.37     & 99.54           & \textbf{100.00}  & \textbf{100.00}       & \textbf{100.00}   & 100            & \textbf{100.00} & \textbf{100.00}    \\
Metal Nut         & 96.10      & 96.53                 & {\ul 97.70}             & 96.29          & 94.98     & 96.90           & \textbf{100.00}  & \textbf{100.00}       & \textbf{100.00}   & 99.10          & 99.26           & \textbf{100.00}    \\
Pill              & 88.80      & 91.79                 & {\ul 93.48}             & 96.61          & 94.81     & 97.10           & 97.68         & 97.90              & 98.04          & \textbf{98.60}          & 96.82           & 96.60           \\
Screw             & 96.30      & 96.21                 & {\ul \textbf{98.93}}    & 96.69          & 94.84     & 93.90           & 91.86         & 94.40              & 94.28          & 97.60          & 91.89           & 98.10           \\
Tile              & 99.40      & {\ul \textbf{100.00}} & {\ul \textbf{100.00}}   & 94.13          & 89.41     & 94.96           & 98.63         & 98.77              & 98.48          & \textbf{100.00}   & 99.88           & 98.70           \\
Toothbrush        & 98.60      & {\ul \textbf{100.00}} & {\ul \textbf{100.00}}   & 97.48          & 97.48     & 97.16           & 98.06         & 98.06              & 98.33          & 91.90          & 99.65           & \textbf{100.00}    \\
Transistor        & 91.10      & {\ul 94.08}           & 93.92                   & 97.03          & 97.74     & 96.95           & 96.88         & 96.83              & 96.88          & 99.30          & 95.21           & \textbf{100.00}    \\
Wood              & 99.80      & 99.83                 & {\ul \textbf{100.00}}   & 94.92          & 94.13     & 95.67           & 99.39         & 99.47              & 99.65          & \textbf{100.00}   & 99.12           & 99.20           \\
Zipper            & 95.10      & {\ul 96.30}           & 95.88                   & 98.68          & 97.36     & 98.69           & 88.87         & 92.31              & 88.05          & \textbf{99.70} & 98.48           & 98.80           \\ \hline
Avg AUROC         & 94.69      & 96.67                 & {\ul 96.97}             & 97.03        & 95.51     & 97.10           & 97.80         & 98.21              & 97.91          & 98.72          & 98.26             & \textbf{99.06}  \\ \hline
\end{tabular}%
}
\caption{Comparison of area under ROC results in \% on MVTec AD dataset. Bold font indicates the best category result, while underlined values show the best result between DifferNet variations. RN18 refers to the network backbone, ResNet-18.}
\label{tab:mvtecad}
\end{table*}

Additionally, AttentDifferNet (SENet) demonstrates notable performance improvements compared to the standard DifferNet, achieving the highest AUROC scores in 14 out of 15 categories. Regarding other attention-based variations, FastFlow (CBAM-RN18) maintains its trend of minor improvements over vanilla FastFlow (RN18) as shown in section~\ref{sec:inspladcomp}, evident from its AUROC score of 98.50\% in the Capsule category. Meanwhile, attention-based RD++ methods (SE-RN18 and CBAM-RN18) show incremental yet noteworthy enhancements. Note that RD++ (SE-RN18) reaches top-3 overall performance in MVTec-AD only behind CFLOW-AD and PatchCore with their original, much deeper backbones than SE-RN18.

When considering the overall performance among DifferNets, AttentDifferNet (CBAM) achieves the highest at 96.97\%, surpassing DifferNet's average AUROC of 94.69\%. Its performance in the Screw category is a highlight, reaching state-of-the-art. This again emphasizes the improvement obtained by incorporating attention modules, even though controlled environments are not the target.

\subsubsection{Semiconductor Wafer Dataset}

Table~\ref{tab:semiconwafer} displays the results in the Semiconductor Wafer dataset. Within the street classification category, AttentDifferNet (SENet) achieves a 90.44\% AUROC, surpassing DifferNet and AttentDifferNet (CBAM) by at least four percentage points. However, it is important to acknowledge that CS-Flow achieves an AUROC score of 97.19\%, being the best in this category by seven percentage points.

Moving to the Chip Classification category, AttentDifferNet (CBAM) stands out with the AUROC score of 93.39\%, outperforming both DifferNet with 91.09\% and AttentDifferNet (SENet) with 89.96\%. Notably, PatchCore achieves a noteworthy AUROC score of 93.90\%, indicating its strong performance within this category.

Despite not achieving the best results, AttentDifferNet consistently improves standard DifferNet results, showing that using attention blocks is also beneficial in this domain.

\begin{table*}[ht!]
  \centering
\resizebox{\textwidth}{!}{%
\begin{tabular}{lccccccc}
\toprule
Category        & DifferNet  & \makecell{AttentDifferNet-SE \\\textbf{(Ours)}} & \makecell{AttentDifferNet-CBAM \\\textbf{(Ours)}}  & \makecell{FastFlow \\ (RN18)} & CS-Flow         & CFLOW-AD  & PatchCore      \\
\midrule                                                                                                                                                                                        
Street          & 86.40      & \underline{90.44}                               & 84.53                                              & 80.94    & \textbf{97.19}  & 70.86     & 79.26          \\
Chip            & 91.09      & 89.96                                           & \underline{93.39}                                  & 76.27    & 90.31           & 92.01     & \textbf{93.90} \\ \midrule
Average AUROC   & 88.74      & \underline{90.18}                               & 88.96                                              & 78.60    & \textbf{93.75}  & 81.44     & 86.58          \\ \bottomrule
\end{tabular}
}
  \caption{Comparison of area under ROC results in \% on Semiconductor Wafer dataset. Bold font indicates the best category result, while underlined values show the best result between DifferNet variations.}
\label{tab:semiconwafer}
\end{table*}

\subsection{Qualitative Results}

\begin{figure*}
    \centering
    \setlength{\tabcolsep}{1pt}
    \begin{tabular}{ccccc}
        \multicolumn{5}{c}{{Input images}} \\
        \noalign{\vskip 2pt}
        \includegraphics[width=0.28\columnwidth]{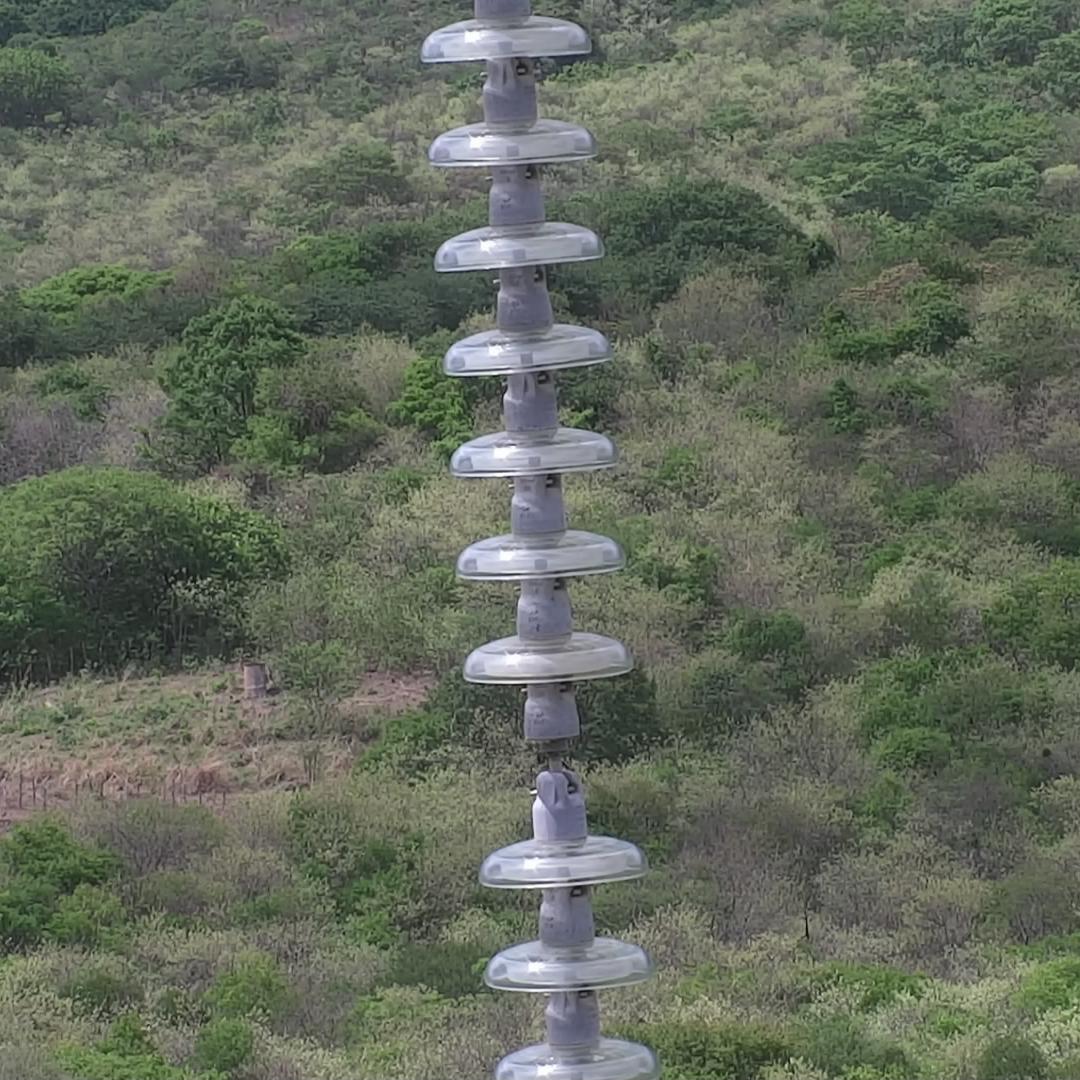}  &
        \includegraphics[width=0.28\columnwidth]{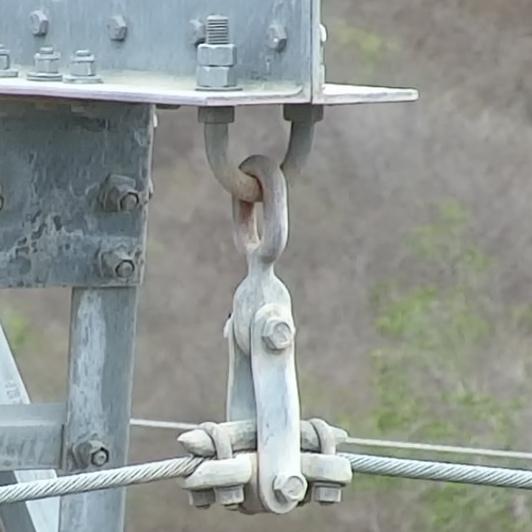} &
        \includegraphics[width=0.28\columnwidth]{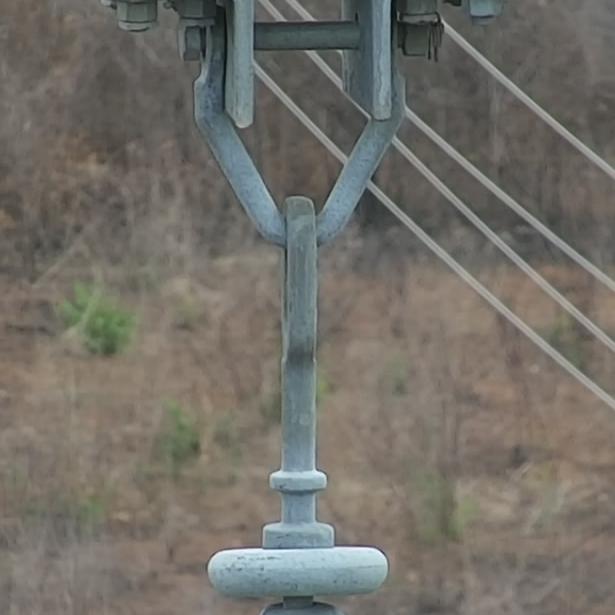}  &
        \includegraphics[width=0.28\columnwidth]{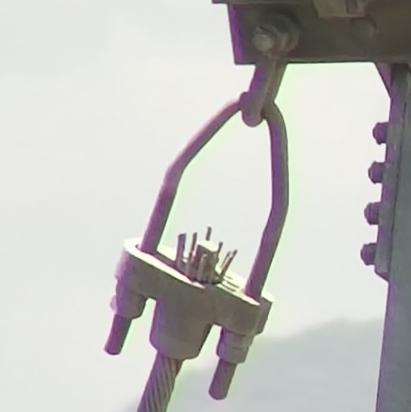} &
        \includegraphics[width=0.28\columnwidth]{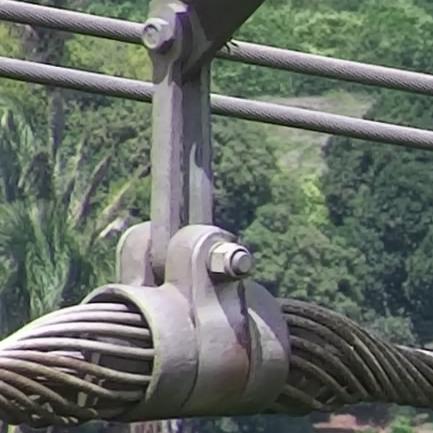} \\
        \noalign{\vskip 4pt}
        \multicolumn{5}{c}{{DifferNet}} \\
        \noalign{\vskip 2pt}
        \includegraphics[width=0.28\columnwidth]{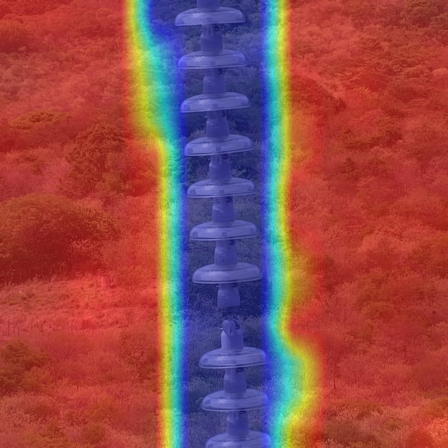}  &
        \includegraphics[width=0.28\columnwidth]{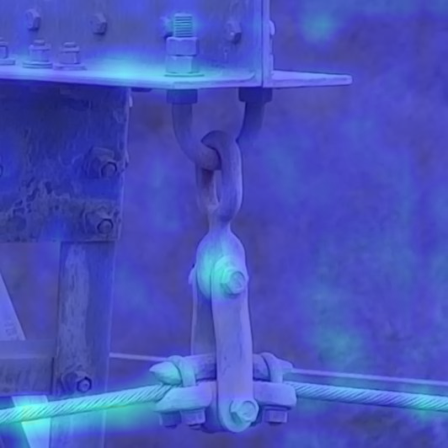} &
        \includegraphics[width=0.28\columnwidth]{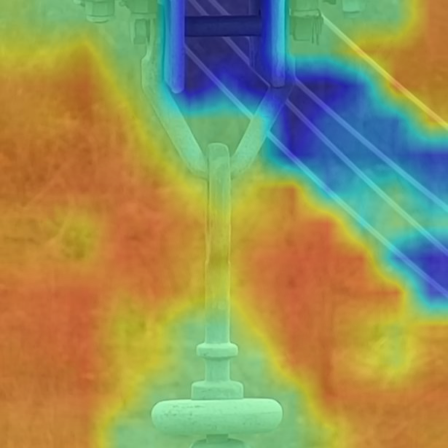}  &
        \includegraphics[width=0.28\columnwidth]{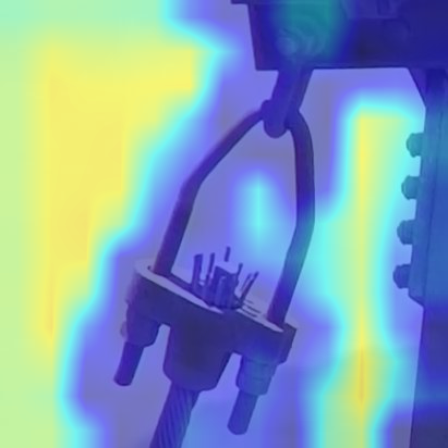} &
        \includegraphics[width=0.28\columnwidth]{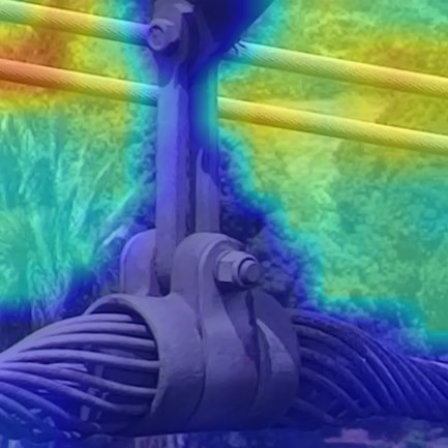} \\
        \noalign{\vskip 4pt}
        \multicolumn{5}{c}{AttentDifferNet \textbf{(Ours)}} \\
        \noalign{\vskip 2pt}
        \includegraphics[width=0.28\columnwidth]{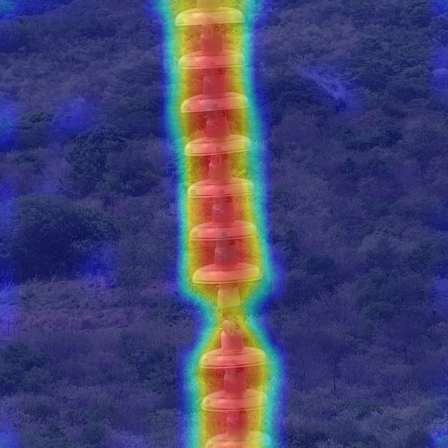}  &
        \includegraphics[width=0.28\columnwidth]{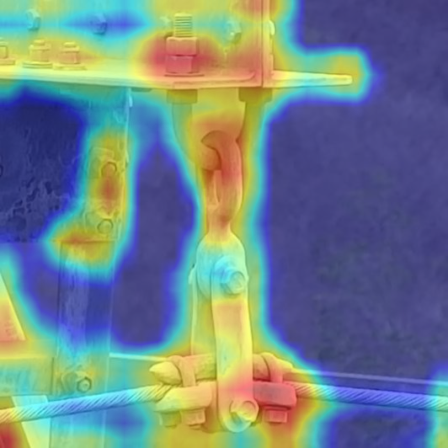} &
        \includegraphics[width=0.28\columnwidth]{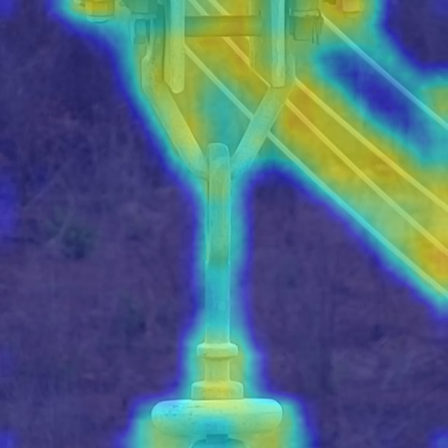}  &
        \includegraphics[width=0.28\columnwidth]{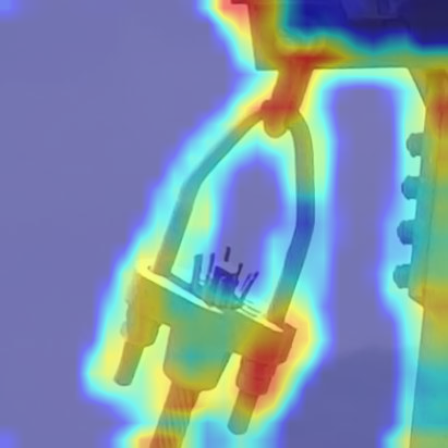} &
        \includegraphics[width=0.28\columnwidth]{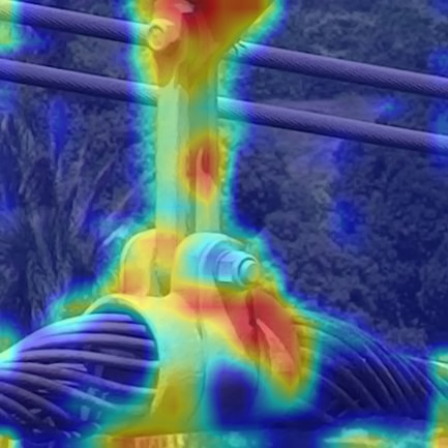} \\
    \end{tabular}

    \caption{Exemplary Grad-CAM-based class activation mapping comparison (blue to red scale, where blue means lower activation and red means higher activation) for DifferNet's backbone versus AttentDifferNet (SENet)'s backbone in all five categories from InsPLAD-fault: Glass Insulator, Lightning Rod Suspension, Polymer Insulator Upper Shackle, and Vari-grip, respectively.}
    \label{fig:inspladquali}
\end{figure*}

\begin{figure*}
    \centering
    \setlength{\tabcolsep}{1pt}
    \begin{tabular}{ccccccc}
        \multicolumn{7}{c}{Input images} \\
        \noalign{\vskip 2pt}
        \includegraphics[width=0.18\columnwidth]{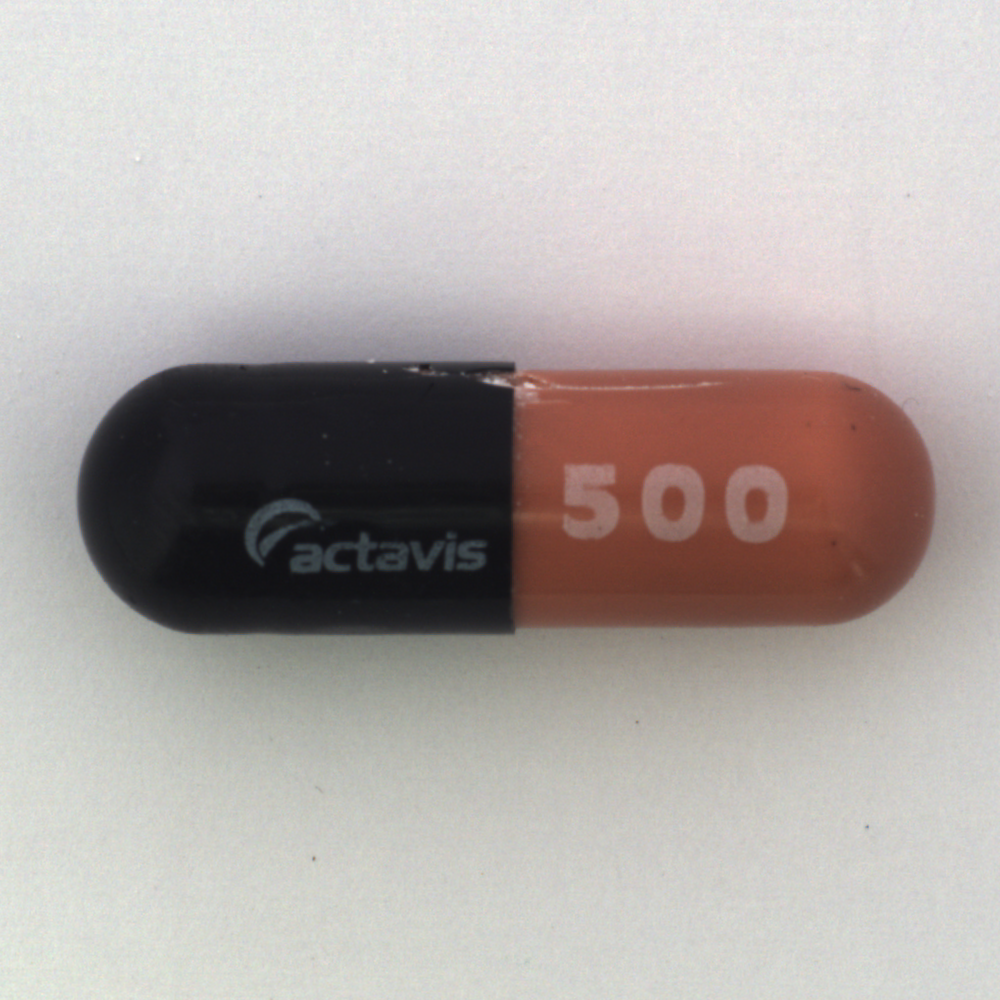} &
        \includegraphics[width=0.18\columnwidth]{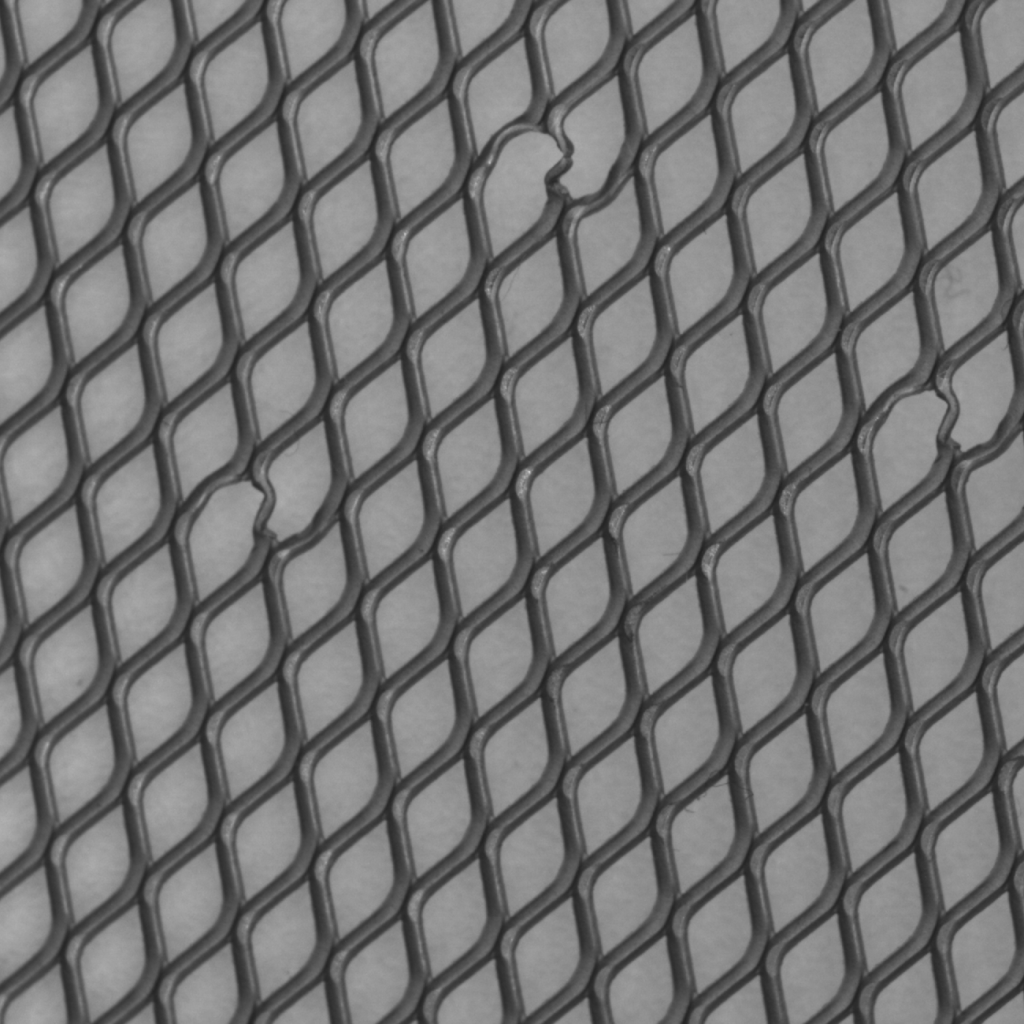} &
        \includegraphics[width=0.18\columnwidth]{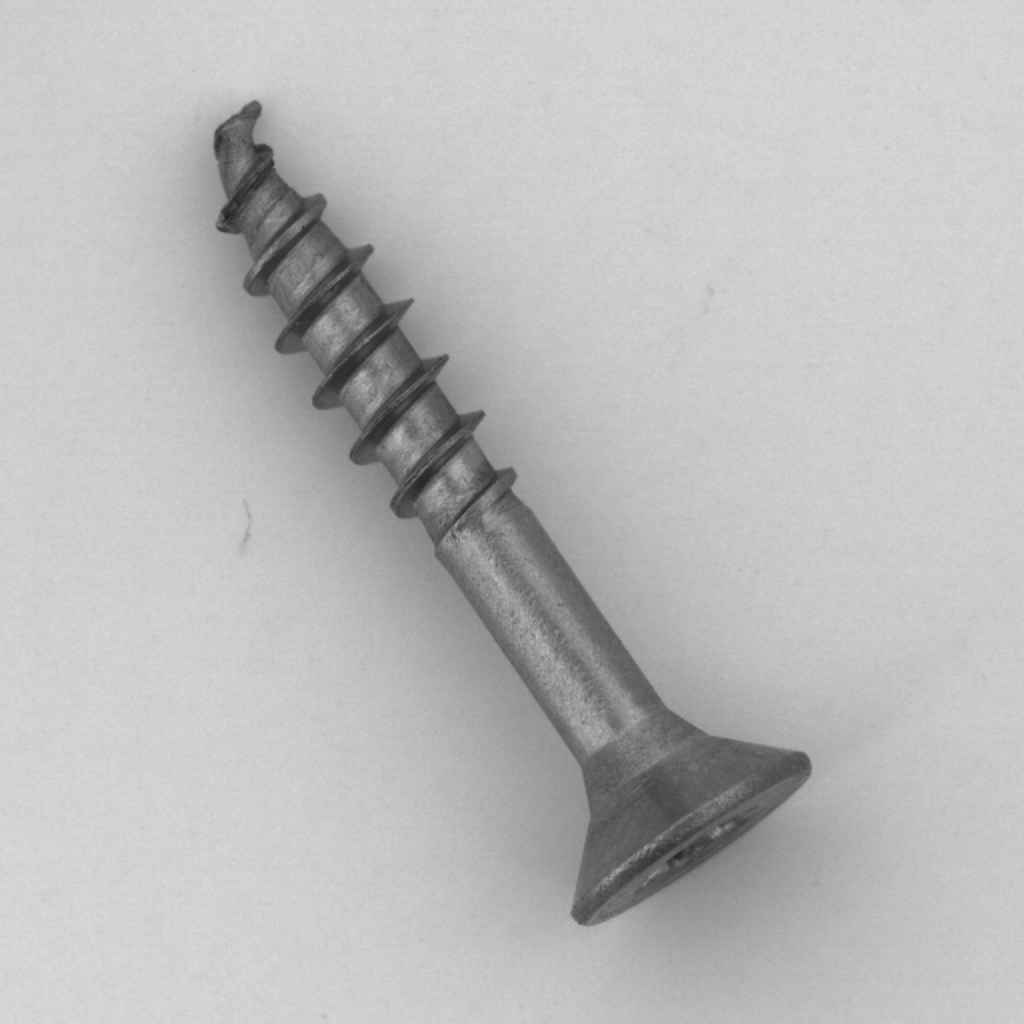} &
        \includegraphics[width=0.18\columnwidth]{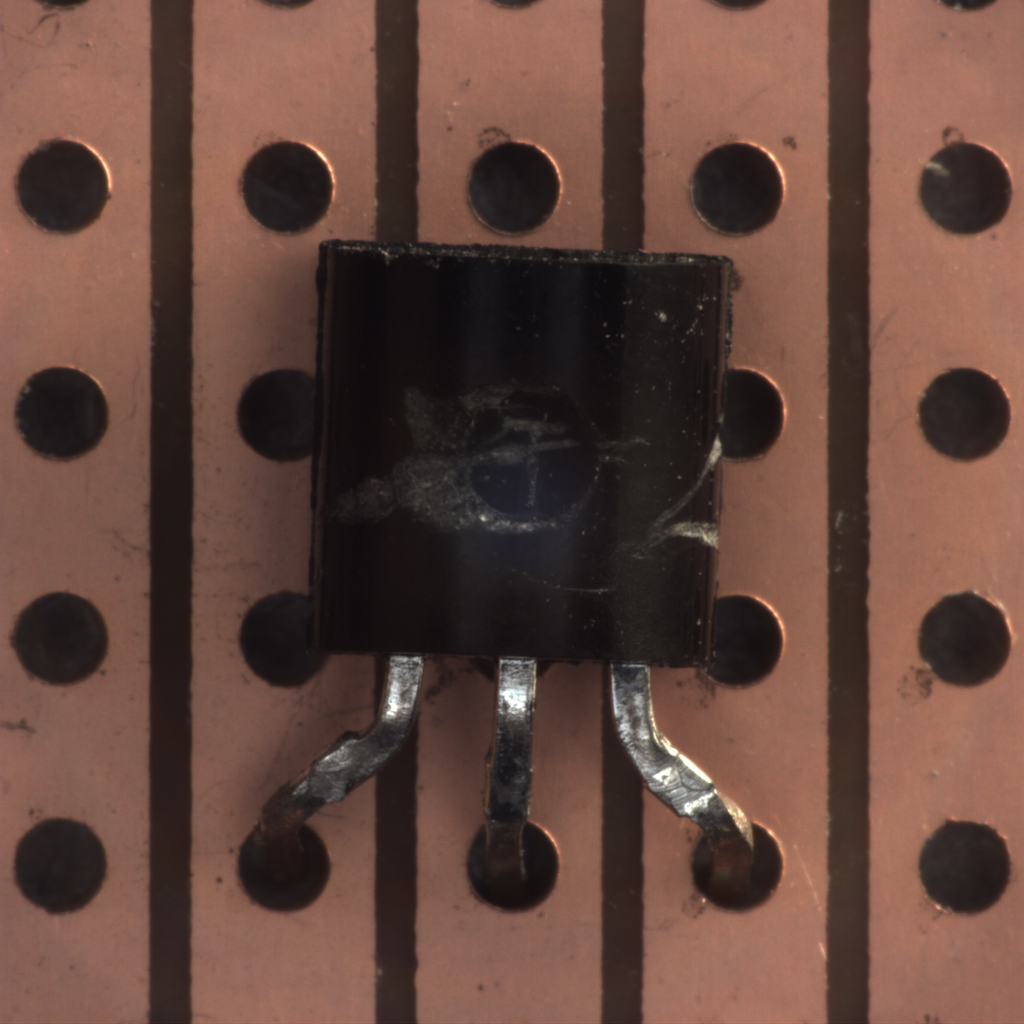} &
        \includegraphics[width=0.18\columnwidth]{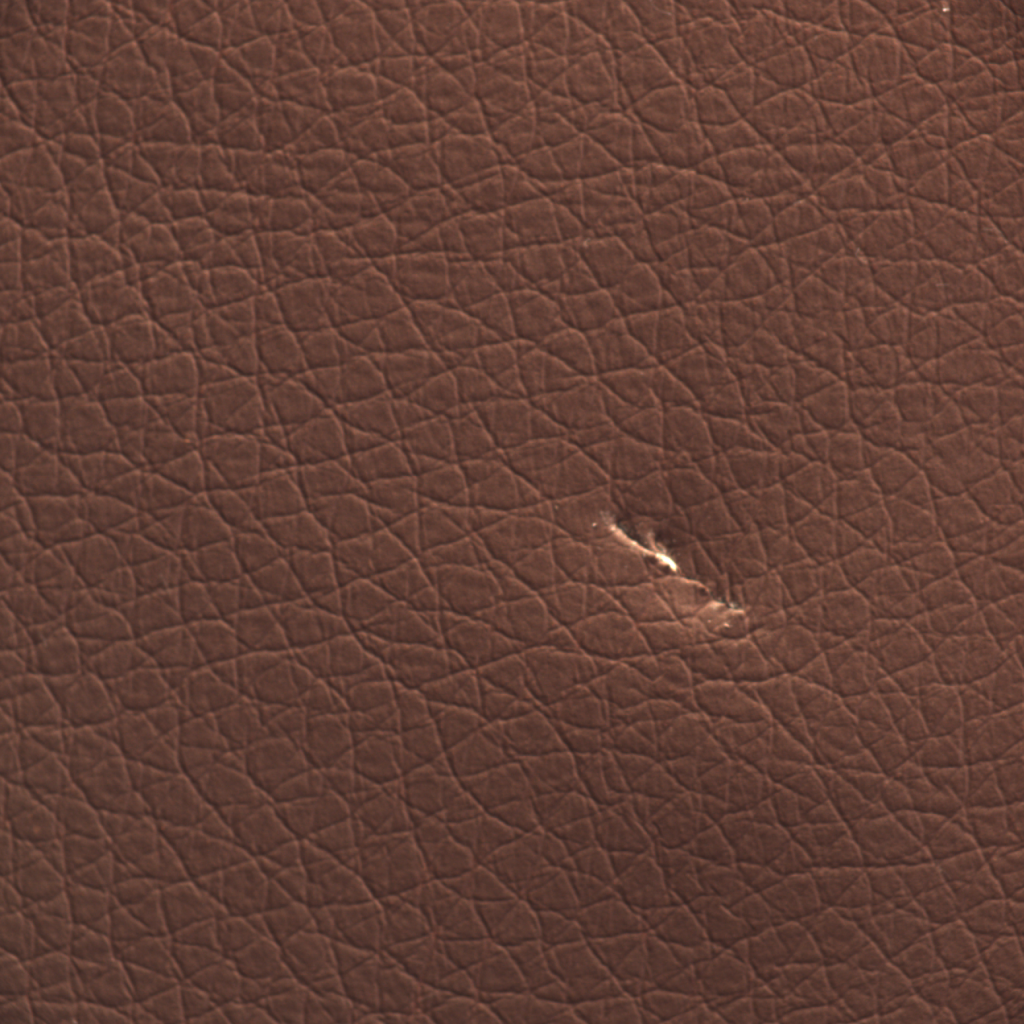} &
        \includegraphics[width=0.18\columnwidth]{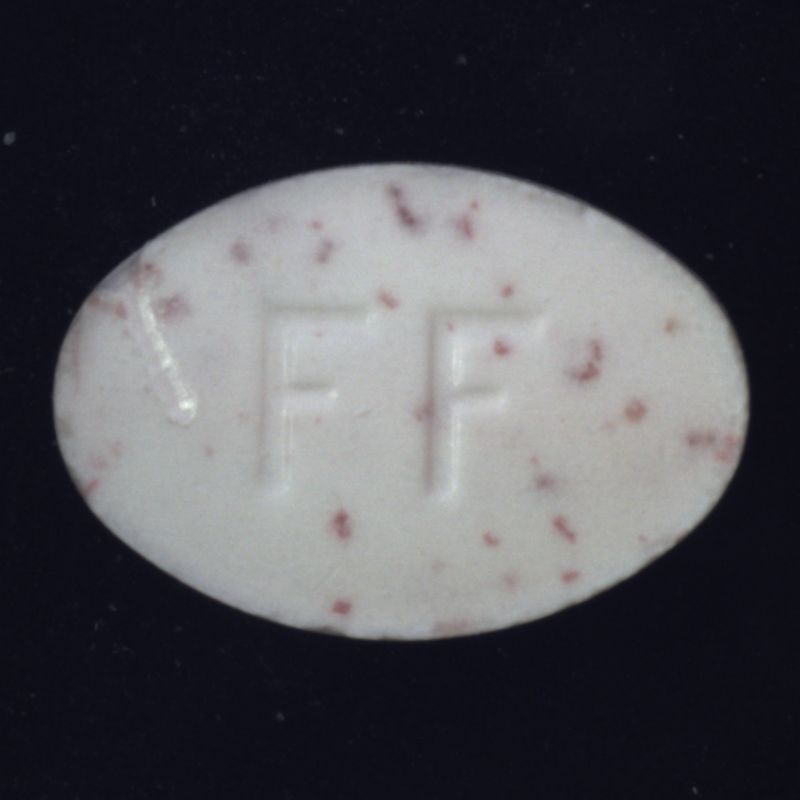} &
        \includegraphics[width=0.18\columnwidth]{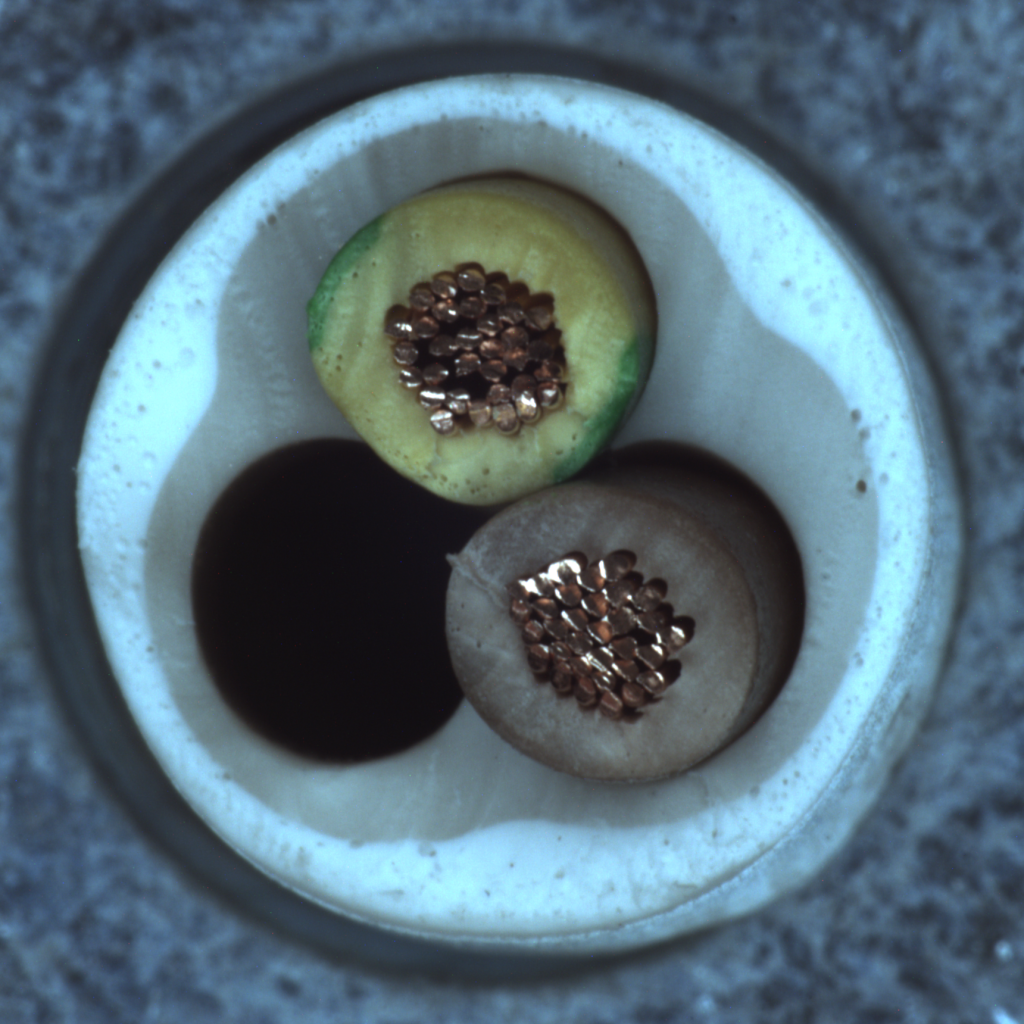}
        \\
        \noalign{\vskip 4pt}
        \multicolumn{7}{c}{DifferNet} \\
        \noalign{\vskip 2pt}
        \includegraphics[width=0.18\columnwidth]{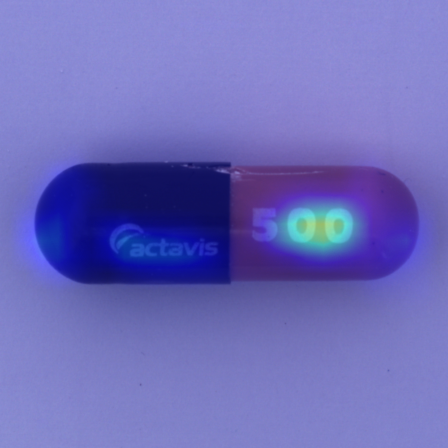} &
        \includegraphics[width=0.18\columnwidth]{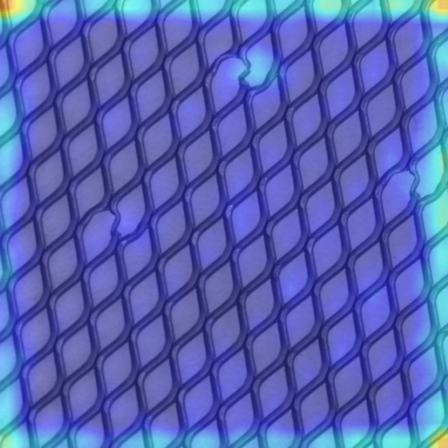} &
        \includegraphics[width=0.18\columnwidth]{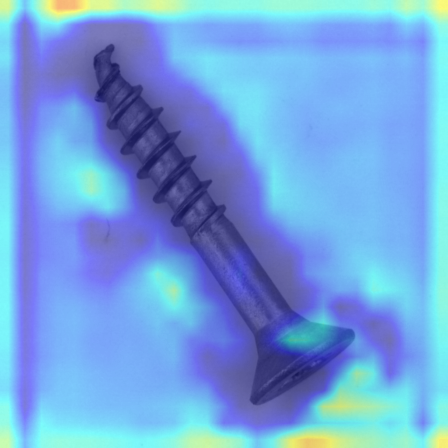} &
        \includegraphics[width=0.18\columnwidth]{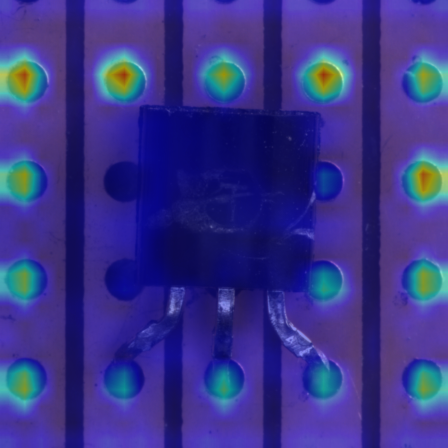} &
        \includegraphics[width=0.18\columnwidth]{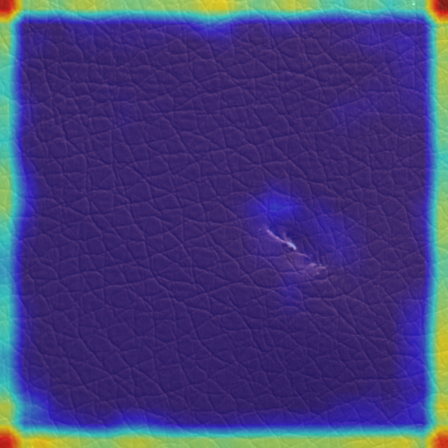} &
        \includegraphics[width=0.18\columnwidth]{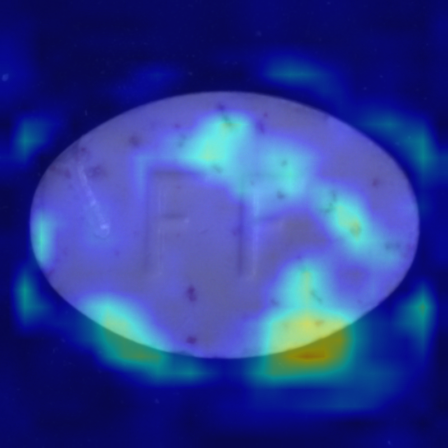} &
        \includegraphics[width=0.18\columnwidth]{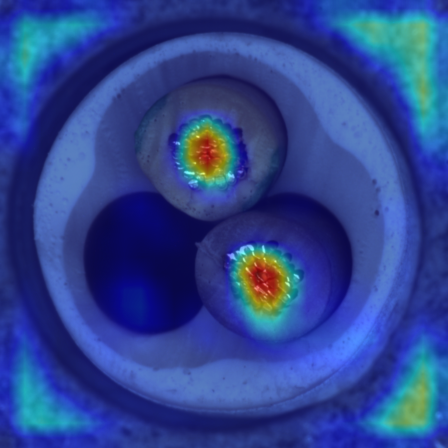}
        \\
        \noalign{\vskip 4pt}
        \multicolumn{7}{c}{AttentDifferNet \textbf{(Ours)}} \\
        \noalign{\vskip 2pt}
        \includegraphics[width=0.18\columnwidth]{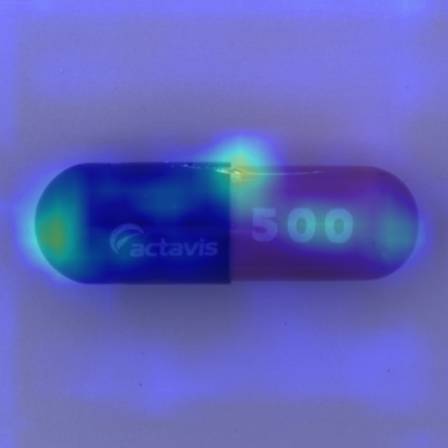} &
        \includegraphics[width=0.18\columnwidth]{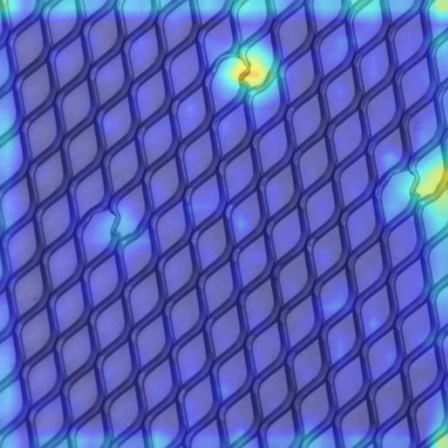} &
        \includegraphics[width=0.18\columnwidth]{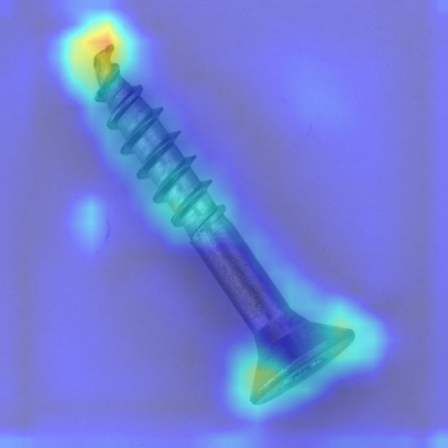} &
        \includegraphics[width=0.18\columnwidth]{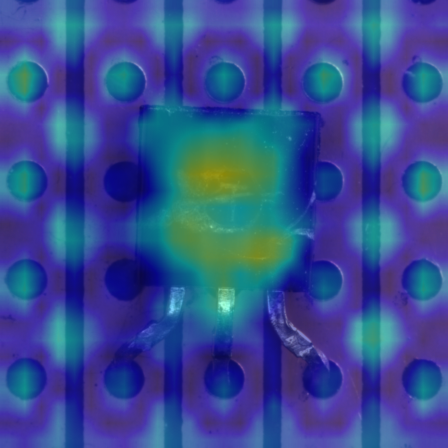} &
        \includegraphics[width=0.18\columnwidth]{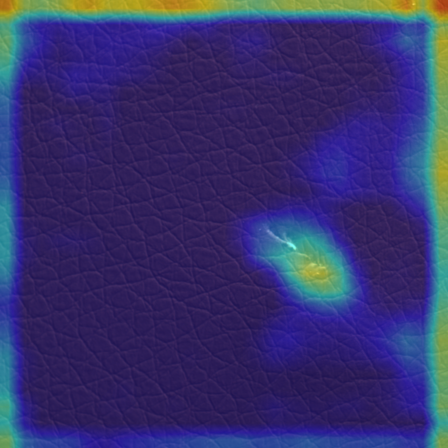} &
        \includegraphics[width=0.18\columnwidth]{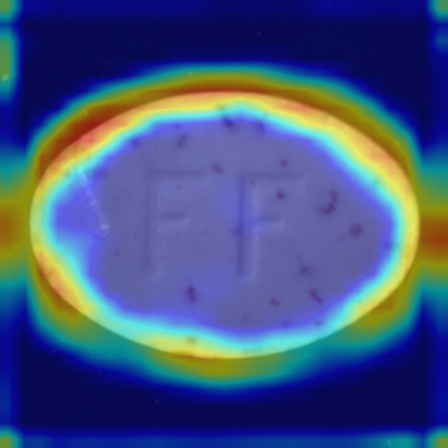} &
        \includegraphics[width=0.18\columnwidth]{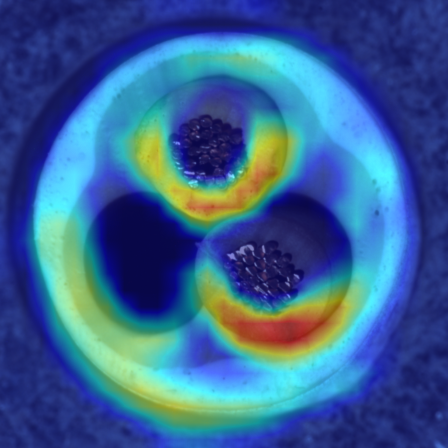} 
        \\
    \end{tabular}

    \caption{Exemplary Grad-CAM-based class activation mapping comparison for DifferNet's backbone versus AttentDifferNet (SENet)'s backbone given seven categories from MVTec AD from left to right: Capsule, Grid, Screw, Transistor, Leather, Pill, and Cable.
    }
    \label{fig:mvtecquali}
\end{figure*}

The qualitative results use the Grad-CAM tool \cite{jacobgilpytorchcam,selvaraju2017gradcam} as an explainable AI tool to reveal where the network is focusing on to make its decisions. We compare the feature extractors of DifferNet and AttentDifferNet, considering all categories from InsPLAD (Figure~\ref{fig:inspladquali}) and some categories of MVTec-AD (Figure~\ref{fig:mvtecquali}). In Figure~\ref{fig:inspladquali}, AttentDifferNet (SENet) is able to focus on more significant features for all classes, such as the object that was supposed to be analyzed in the foreground, reducing the impact of the background as expected. In the first object (Glass Insulator), the missing cap is now taken into account directly as the backbone focuses on the object. In the middle column (Polymer Insulator Upper Shackle), the network improves its focus on the foreground object but also focuses on the transmission tower wires, which may disturb the model outcome since it is not part of the main object of interest.

The other comparisons are from MVTec AD data. Note that DifferNet's behavior differs from the anomaly localization maps presented in the original DifferNet paper. That is expected since we are focusing on the feature extractor step only. Grad-CAM's target layer is the last convolutional layer of the feature extractor network, where it has the best compromise between high-level semantics and detailed spatial information, as the Grad-CAM authors \cite{selvaraju2017gradcam} recommend.
Considering AttentDifferNet, it was even more specific on MVTec AD, apparently focusing on the anomaly/defect itself, both in objects and texture categories.

\subsection{Ablation}

We performed a brief ablation study to assess the impact of different combinations of attention blocks added to the model architecture. It consisted of training the proposed AttentDifferNet by removing the attention blocks AB\textsubscript{1}, AB\textsubscript{2}, and AB\textsubscript{3} (shown in Figure~\ref{fig:attentdiffernet}) from the model architecture in all possible combinations. Table~\ref{tab:ablation} shows the study results using the Glass Insulator object from the InsPLAD-fault dataset. Note how using only one attention block instead of none does not yield better results, but using two attention blocks instead of one results in improved results on average. It is also noteworthy that placing the attention block in later layers appears to cause a greater impact on AUROC. 

\begin{table}[ht]
    \centering
\resizebox{.55\columnwidth}{!}{%
    \begin{tabular}{cccc}
    \toprule
    AB\textsubscript{1} & AB\textsubscript{2} & AB\textsubscript{3} & AUROC [\%] \\
    \midrule
    \ding{55}  & \ding{55} & \ding{55} & 82.81  \\
    \checkmark  & \ding{55} & \ding{55} & 80.85  \\
    \ding{55} & \checkmark & \ding{55} & 81.64  \\
    \ding{55} & \ding{55} & \checkmark  & 82.85  \\
    \checkmark  & \checkmark  & \ding{55} &  83.17 \\
    \checkmark  & \ding{55} & \checkmark  & 82.32  \\
    \ding{55}  & \checkmark  & \checkmark  &  84.91 \\
    \checkmark  & \checkmark  & \checkmark  & 86.57  \\
    \bottomrule
    \end{tabular}%
}
    \caption{Area under ROC results in \% for our ablation study showing different combinations of attention blocks considering the Glass Insulator category from InsPLAD-fault.}
    \label{tab:ablation}
\end{table}

\section{Conclusions and Outlook}

The main hypothesis of this work was that attention modules help to improve the performance of state-of-the-art anomaly detection methods in an in-the-wild/uncontrolled environment scenario. We proposed AttentDifferNet, an unsupervised anomaly detection method based on distribution mappings through normalizing flows that benefits from attention mechanisms by strategically coupling modular attention blocks to its feature extraction step. AttentDifferNet achieves image-level state-of-the-art performance on InsPLAD-fault, an anomaly detection in-the-wild dataset. We also experimented with attention-based versions of two state-of-the-art anomaly detection methods in a similar fashion to DifferNet, FastFlow and RD++. Generally, the attention-based versions of FastFlow and RD++ present an increase in performance compared to their standard versions. We also show that AttentDifferNet not only maintains the model performance compared to DifferNet in controlled environments, but can also improve in virtually all categories of two relevant controlled environments' datasets for anomaly detection: the MVTec AD and the Semiconductor Wafer dataset.
Our qualitative analysis supports the improved quantitative performance, showing that the proposed network can focus on the object to be inspected when in the wild. In a controlled environment, it focuses on the defect.

This work implies that state-of-the-art unsupervised anomaly detection methods have limitations in uncontrolled/in-the-wild environments. It also portrays how the usage of attention blocks is well-suited to deal with such limitations and their potential to improve anomaly detection analysis at a pixel level with only image-level annotations.

\section*{Acknowledgements}

This research was co-funded by the German Academic Exchange Service (DAAD) and the Coordenação de Aperfeiçoamento de Pessoal de Nível Superior (CAPES) within the funding program ``Co-financed Short-Term Research Grant Brazil, 2022 (ID: 57594818)''.

{\small
\bibliographystyle{ieee_fullname}
\bibliography{egbib, library_Tobias}

\begin{thebibliography}{10}\itemsep=-1pt

\bibitem{akcay2019skipganomaly}
Samet Akçay, Amir Atapour-Abarghouei, and Toby~P. Breckon.
\newblock {Skip-GANomaly: Skip Connected and Adversarially Trained
  Encoder-Decoder Anomaly Detection}.
\newblock In {\em 2019 International Joint Conference on Neural Networks
  (IJCNN)}, pages 1--8, 2019.

\bibitem{bergmann2021mvtec}
Paul Bergmann, Kilian Batzner, Michael Fauser, David Sattlegger, and Carsten
  Steger.
\newblock {The MVTec Anomaly Detection Dataset: A Comprehensive Real-world
  Dataset for Unsupervised Anomaly Detection}.
\newblock {\em International Journal of Computer Vision}, 129(4):1038--1059,
  2021.

\bibitem{bergmann2022Beyond}
Paul Bergmann, Kilian Batzner, Michael Fauser, David Sattlegger, and Carsten
  Steger.
\newblock Beyond dents and scratches: Logical constraints in unsupervised
  anomaly detection and localization.
\newblock {\em Int. J. Comput. Vision}, 130(4):947–969, apr 2022.

\bibitem{bergmann2019mvtec}
Paul Bergmann, Michael Fauser, David Sattlegger, and Carsten Steger.
\newblock {MVTec AD -- A Comprehensive Real-World Dataset for Unsupervised
  Anomaly Detection}.
\newblock In {\em Proceedings of the IEEE/CVF Conference on Computer Vision and
  Pattern Recognition (CVPR)}, June 2019.

\bibitem{bhattacharya2021interleaved}
Gaurab Bhattacharya, Bappaditya Mandal, and Niladri~B. Puhan.
\newblock {Interleaved Deep Artifacts-Aware Attention Mechanism for Concrete
  Structural Defect Classification}.
\newblock {\em IEEE Transactions on Image Processing}, 30:6957--6969, 2021.

\bibitem{Cheon2019}
Sejune Cheon, Hankang Lee, Chang~Ouk Kim, and Seok~Hyung Lee.
\newblock {Convolutional neural network for wafer surface defect classification
  and the detection of unknown defect class}.
\newblock {\em IEEE Transactions on Semiconductor Manufacturing},
  32(2):163--170, 2019.

\bibitem{cohen2020spade}
Niv Cohen and Yedid Hoshen.
\newblock {Sub-Image Anomaly Detection with Deep Pyramid Correspondences},
  2020.

\bibitem{defard2021padim}
Thomas Defard, Aleksandr Setkov, Angelique Loesch, and Romaric Audigier.
\newblock {PaDiM: A Patch Distribution Modeling Framework for Anomaly Detection
  and Localization}.
\newblock In Alberto Del~Bimbo, Rita Cucchiara, Stan Sclaroff, Giovanni~Maria
  Farinella, Tao Mei, Marco Bertini, Hugo~Jair Escalante, and Roberto Vezzani,
  editors, {\em Pattern Recognition. ICPR International Workshops and
  Challenges}, pages 475--489, Cham, 2021. Springer International Publishing.

\bibitem{Chen2000}
{Fei-Long Chen} and {Shu-Fan Liu}.
\newblock {A neural-network approach to recognize defect spatial pattern in
  semiconductor fabrication}.
\newblock {\em IEEE Transactions on Semiconductor Manufacturing},
  13(3):366--373, 2000.

\bibitem{Friedrich2023}
Michael Friedrich, Tobias Schlosser, and Danny Kowerko.
\newblock {Simulation of Semiconductor Wafer Dicing Induced Faults on Chips and
  their Application as Augmentation Method for a Deep Learning Based Visual
  Inspection System}.
\newblock {\em Journal of Intelligent Manufacturing (Preprint)}, 2023.

\bibitem{ge2021anomaly}
Bangbang Ge, Chunping Hou, Yang Liu, Zhipeng Wang, and Ruiheng Wu.
\newblock {Anomaly Detection of Power Line Insulator from Aerial Imagery with
  Attribute Self-supervised Learning}.
\newblock {\em International Journal of Remote Sensing}, 42(23):8819--8839,
  2021.

\bibitem{jacobgilpytorchcam}
Jacob Gildenblat and contributors.
\newblock {PyTorch library for CAM methods}.
\newblock \url{https://github.com/jacobgil/pytorch-grad-cam}, 2021.

\bibitem{gudovskiy2022cflowad}
Denis Gudovskiy, Shun Ishizaka, and Kazuki Kozuka.
\newblock {CFLOW-AD: Real-Time Unsupervised Anomaly Detection With Localization
  via Conditional Normalizing Flows}.
\newblock In {\em Proceedings of the IEEE/CVF Winter Conference on Applications
  of Computer Vision (WACV)}, pages 98--107, January 2022.

\bibitem{han2021madgan}
Changhee Han, Leonardo Rundo, Kohei Murao, Tomoyuki Noguchi, Yuki Shimahara,
  Zolt{\'a}n~{\'A}d{\'a}m Milacski, Saori Koshino, Evis Sala, Hideki Nakayama,
  and Shin’ichi Satoh.
\newblock {MADGAN: Unsupervised Medical Anomaly Detection GAN Using Multiple
  Adjacent Brain MRI Slice Reconstruction}.
\newblock {\em BMC bioinformatics}, 22(2):1--20, 2021.

\bibitem{hu2018senet}
Jie Hu, Li Shen, and Gang Sun.
\newblock {Squeeze-and-Excitation Networks}.
\newblock In {\em Proceedings of the IEEE Conference on Computer Vision and
  Pattern Recognition (CVPR)}, June 2018.

\bibitem{Huang2007}
Chenn-Jung Huang.
\newblock {Clustered defect detection of high quality chips using
  self-supervised multilayer perceptron}.
\newblock {\em Expert Systems with Applications}, 33(4):996--1003, 2007.

\bibitem{Huang2015}
Szu-Hao Huang and Ying-Cheng Pan.
\newblock {Automated visual inspection in the semiconductor industry: A
  survey}.
\newblock {\em Computers in Industry}, 66, 2015.

\bibitem{huang2020surface}
Yibin Huang, Congying Qiu, and Kui Yuan.
\newblock {Surface Defect Saliency of Magnetic Tile}.
\newblock {\em The Visual Computer}, 36:85--96, 2020.

\bibitem{krizhevsky2017alexnet}
Alex Krizhevsky, Ilya Sutskever, and Geoffrey~E. Hinton.
\newblock {ImageNet Classification with Deep Convolutional Neural Networks}.
\newblock {\em Commun. ACM}, 60(6):84–90, may 2017.

\bibitem{liu2023deep}
Jiaqi Liu, Guoyang Xie, Jingbao Wang, Shangnian Li, Chengjie Wang, Feng Zheng,
  and Yaochu Jin.
\newblock {Deep Industrial Image Anomaly Detection: A Survey}.
\newblock {\em arXiv e-prints}, pages arXiv--2301, 2023.

\bibitem{liu2002wafer}
S~F Liu, F~L Chen, and W~B Lu.
\newblock {Wafer bin map recognition using a neural network approach}.
\newblock {\em International Journal of production research},
  40(10):2207--2223, 2002.

\bibitem{Nakazawa2018}
Takeshi Nakazawa and Deepak~V. Kulkarni.
\newblock {Wafer map defect pattern classification and image retrieval using
  convolutional neural network}.
\newblock {\em IEEE Transactions on Semiconductor Manufacturing},
  31(2):309--314, 2018.

\bibitem{OLeary2020}
Jared O'Leary, Kapil Sawlani, and Ali Mesbah.
\newblock {Deep learning for classification of the chemical composition of
  particle defects on semiconductor wafers}.
\newblock {\em IEEE Transactions on Semiconductor Manufacturing}, 33(1):72--85,
  2020.

\bibitem{ristea2022selfsupervised}
Nicolae-C\u{a}t\u{a}lin Ristea, Neelu Madan, Radu~Tudor Ionescu, Kamal
  Nasrollahi, Fahad~Shahbaz Khan, Thomas~B. Moeslund, and Mubarak Shah.
\newblock {Self-Supervised Predictive Convolutional Attentive Block for Anomaly
  Detection}.
\newblock In {\em Proceedings of the IEEE/CVF Conference on Computer Vision and
  Pattern Recognition (CVPR)}, pages 13576--13586, June 2022.

\bibitem{roth2022patchcore}
Karsten Roth, Latha Pemula, Joaquin Zepeda, Bernhard Sch\"olkopf, Thomas Brox,
  and Peter Gehler.
\newblock {Towards Total Recall in Industrial Anomaly Detection}.
\newblock In {\em Proceedings of the IEEE/CVF Conference on Computer Vision and
  Pattern Recognition (CVPR)}, pages 14318--14328, June 2022.

\bibitem{rudolph2021differnet}
Marco Rudolph, Bastian Wandt, and Bodo Rosenhahn.
\newblock {Same Same but {DifferNet}: Semi-Supervised Defect Detection With
  Normalizing Flows}.
\newblock In {\em Proceedings of the IEEE/CVF Winter Conference on Applications
  of Computer Vision (WACV)}, pages 1907--1916, January 2021.

\bibitem{rudolph2022csflow}
Marco Rudolph, Tom Wehrbein, Bodo Rosenhahn, and Bastian Wandt.
\newblock {Fully Convolutional Cross-Scale-Flows for Image-Based Defect
  Detection}.
\newblock In {\em Proceedings of the IEEE/CVF Winter Conference on Applications
  of Computer Vision (WACV)}, pages 1088--1097, January 2022.

\bibitem{schlegl2019fanogan}
Thomas Schlegl, Philipp Seeböck, Sebastian~M. Waldstein, Georg Langs, and
  Ursula Schmidt-Erfurth.
\newblock {f-AnoGAN: Fast Unsupervised Anomaly Detection with Generative
  Adversarial Networks}.
\newblock {\em Medical Image Analysis}, 54:30--44, 2019.

\bibitem{schlosser2022improving}
Tobias Schlosser, Michael Friedrich, Frederik Beuth, and Danny Kowerko.
\newblock {Improving Automated Visual Fault Inspection for Semiconductor
  Manufacturing Using a Hybrid Multistage System of Deep Neural Networks}.
\newblock {\em Journal of Intelligent Manufacturing}, 33(4):1099--1123, 2022.

\bibitem{selvaraju2017gradcam}
Ramprasaath~R. Selvaraju, Michael Cogswell, Abhishek Das, Ramakrishna Vedantam,
  Devi Parikh, and Dhruv Batra.
\newblock {Grad-CAM: Visual Explanations From Deep Networks via Gradient-Based
  Localization}.
\newblock In {\em Proceedings of the IEEE International Conference on Computer
  Vision (ICCV)}, Oct 2017.

\bibitem{song2023research}
Chuanwang Song, Ziyu Li, Yuming Li, Hao Zhang, Mingjian Jiang, Keyong Hu, and
  Rihong Wang.
\newblock {Research on Blast Furnace Tuyere Image Anomaly Detection, Based on
  the Local Channel Attention Residual Mechanism}.
\newblock {\em Applied Sciences}, 13(2), 2023.

\bibitem{sreenivasan1993automated}
Koduri~K Sreenivasan, Mandyam Srinath, and Alireza Khotanzad.
\newblock {Automated vision system for inspection of IC pads and bonds}.
\newblock {\em IEEE transactions on components, hybrids, and manufacturing
  technology}, 16(3):333--338, 1993.

\bibitem{sun2022scale}
Jian Sun, Hongwei Gao, Xuna Wang, and Jiahui Yu.
\newblock {Scale Enhancement Pyramid Network for Small Object Detection from
  UAV Images}.
\newblock {\em Entropy}, 24(11), 2022.

\bibitem{takimoto2022anomaly}
Hironori Takimoto, Junya Seki, Sulfayanti~F. Situju, and Akihiro Kanagawa.
\newblock {Anomaly Detection Using Siamese Network with Attention Mechanism for
  Few-Shot Learning}.
\newblock {\em Applied Artificial Intelligence}, 36(1):2094885, 2022.

\bibitem{tien2023revisiting}
Tran~Dinh Tien, Anh~Tuan Nguyen, Nguyen~Hoang Tran, Ta~Duc Huy, Soan~T.M.
  Duong, Chanh D.~Tr. Nguyen, and Steven Q.~H. Truong.
\newblock Revisiting reverse distillation for anomaly detection.
\newblock In {\em Proceedings of the IEEE/CVF Conference on Computer Vision and
  Pattern Recognition (CVPR)}, pages 24511--24520, June 2023.

\bibitem{tobin2001integrated}
Kenneth~W Tobin~Jr, Thomas~P Karnowski, and Fred Lakhani.
\newblock {Integrated applications of inspection data in the semiconductor
  manufacturing environment}.
\newblock In {\em Metrology-based Control for Micro-Manufacturing}, volume
  4275, pages 31--40. SPIE, 2001.

\bibitem{vieiraesilva2021stnplad}
André Luiz~Buarque {Vieira-e-Silva}, Heitor de Castro~Felix, Thiago de
  Menezes~Chaves, Francisco Paulo~Magalhães Simões, Veronica Teichrieb,
  Michel~Mozinho dos Santos, Hemir da Cunha~Santiago, Virginia Adélia~Cordeiro
  Sgotti, and Henrique Baptista Duffles Teixeira~Lott Neto.
\newblock {STN PLAD}: A dataset for multi-size power line assets detection in
  high-resolution uav images.
\newblock In {\em 2021 34th SIBGRAPI Conference on Graphics, Patterns and
  Images (SIBGRAPI)}, pages 215--222, 2021.

\bibitem{vieiraesilva2023insplad}
André Luiz~Buarque {Vieira e Silva}, Heitor de Castro~Felix, Francisco
  Simões, Veronica Teichrieb, Michel~Mozinho dos Santos, Hemir Santiago,
  Virginia Sgotti, and Henrique Lott~Neto.
\newblock {InsPLAD: A Dataset and Benchmark for Power Line Asset Inspection in
  UAV Images}.
\newblock {\em International Journal of Remote Sensing}, page (to appear),
  2023.
\newblock Accepted.

\bibitem{wang2020ecanet}
Qilong Wang, Banggu Wu, Pengfei Zhu, Peihua Li, Wangmeng Zuo, and Qinghua Hu.
\newblock Eca-net: Efficient channel attention for deep convolutional neural
  networks.
\newblock In {\em Proceedings of the IEEE/CVF Conference on Computer Vision and
  Pattern Recognition (CVPR)}, June 2020.

\bibitem{woo2018cbam}
Sanghyun Woo, Jongchan Park, Joon-Young Lee, and In~So Kweon.
\newblock {CBAM: Convolutional Block Attention Module}.
\newblock In {\em Proceedings of the European Conference on Computer Vision
  (ECCV)}, September 2018.

\bibitem{Xie2014}
Liangjun Xie, Rui Huang, Nong Gu, and Zhiqiang Cao.
\newblock {A novel defect detection and identification method in optical
  inspection}.
\newblock {\em Neural Computing and Applications}, 24(7-8):1953--1962, 2014.

\bibitem{yan2022cainnflow}
Ruiqing Yan, Fan Zhang, Mengyuan Huang, Wu Liu, Dongyu Hu, Jinfeng Li, Qiang
  Liu, Jinrong Jiang, Qianjin Guo, and Linghan Zheng.
\newblock {CAINNFlow: Convolutional block Attention modules and Invertible
  Neural Networks Flow for anomaly detection and localization tasks}, 2022.

\bibitem{yu2021fastflow}
Jiawei Yu, Ye Zheng, Xiang Wang, Wei Li, Yushuang Wu, Rui Zhao, and Liwei Wu.
\newblock {FastFlow: Unsupervised Anomaly Detection and Localization via 2D
  Normalizing Flows}, 2021.

\bibitem{zhang1999development}
Jiun-Ming Zhang, Ruey-Ming Lin, and Mao-Jiun~J Wang.
\newblock {The development of an automatic post-sawing inspection system using
  computer vision techniques}.
\newblock {\em Computers in Industry}, 40(1):51--60, 1999.

\bibitem{zhang2021sanet}
Qing-Long Zhang and Yu-Bin Yang.
\newblock Sa-net: Shuffle attention for deep convolutional neural networks.
\newblock In {\em ICASSP 2021 - 2021 IEEE International Conference on
  Acoustics, Speech and Signal Processing (ICASSP)}, pages 2235--2239, 2021.

\bibitem{zhang2023industrial}
Zilong Zhang, Zhibin Zhao, Xingwu Zhang, Chuang Sun, and Xuefeng Chen.
\newblock Industrial anomaly detection with domain shift: A real-world dataset
  and masked multi-scale reconstruction, 2023.

\end{thebibliography}
}

\end{document}